\documentclass{article}
\PassOptionsToPackage{numbers, compress}{natbib}

\usepackage[final]{neurips_2025}

\usepackage[utf8]{inputenc} 
\usepackage[T1]{fontenc}    
\usepackage{url}            
\usepackage{booktabs}       
\usepackage{amsfonts}       
\usepackage{nicefrac}       
\usepackage{microtype}      
\usepackage{xcolor}         
\usepackage{enumitem}
\usepackage{xspace}
\usepackage{pifont}
\usepackage{graphicx}
\usepackage{multirow} 
\usepackage{amsmath}
\usepackage{adjustbox}
\usepackage{arydshln}
\usepackage{float}
\usepackage{amssymb}
\usepackage{xspace}
\usepackage{makecell}
\usepackage{wrapfig}
\usepackage{comment}
\usepackage{xcolor}
\usepackage{wrapfig}
\usepackage{float}

\usepackage{algorithm,algorithmicx,algpseudocode}

\usepackage{colortbl}
\usepackage{booktabs}

\newif\ifshowcomments %
\showcommentsfalse     %

\ifshowcomments
  
\else
  
\fi

\makeatletter
\DeclareRobustCommand\onedot{\futurelet\@let@token\@onedot}
\def\@onedot{\ifx\@let@token.\else.\null\fi\xspace}

\def\eg{e.g\onedot}

\makeatother

\usepackage[dvipsnames]{xcolor}
\usepackage{amssymb}
\usepackage{amsmath}
\usepackage{pifont}
\usepackage{multirow}
\usepackage{tabularx}
\usepackage{makecell}
\usepackage{colortbl}
\usepackage{bm}
\usepackage{gensymb}
\usepackage[outline]{contour}
\usepackage[rightcaption]{sidecap}
\usepackage{bbm}
\usepackage{float}
\usepackage[inkscapelatex=false]{svg}

\definecolor{turquoise}{cmyk}{0.65,0,0.1,0.3}
\definecolor{purple}{rgb}{0.65,0,0.65}
\definecolor{dark_green}{rgb}{0, 0.5, 0}
\definecolor{orange}{rgb}{0.8, 0.6, 0.2}
\definecolor{red}{rgb}{0.8, 0.2, 0.2}
\definecolor{darkred}{rgb}{0.6, 0.1, 0.05}
\definecolor{blueish}{rgb}{0.0, 0.3, .6}
\definecolor{light_gray}{rgb}{0.7, 0.7, .7}
\definecolor{pink}{rgb}{1, 0, 1}
\definecolor{greyblue}{rgb}{0.25, 0.25, 1}
\definecolor{LightRed}{rgb}{0.99,0.89,0.89}

\definecolor{orange}{HTML}{FF7F00}

\colorlet{colororange}{orange!45}
\colorlet{colorturquoise}{turquoise!45}
\colorlet{colorgreen}{green!30}

\colorlet{colorFst}{Green!25}       
\colorlet{colorSnd}{SpringGreen!45} 
\colorlet{colorTrd}{Yellow!30}      
\colorlet{colorLow}{darkgray!30}
\newcommand{\fs}{\bf}   
\newcommand{\nd}{\underline}      

\usepackage[pagebackref=true,breaklinks=true,colorlinks,bookmarks=false]{hyperref}

\definecolor{mycitecolor}{HTML}{195a66}
\hypersetup{
  citecolor  = mycitecolor,
  colorlinks = true,
}

\newcommand{\project}{GuideFlow3D\xspace}

\title{\project: Optimization-Guided Rectified Flow For Appearance Transfer}

\author{%
  \begin{minipage}[t]{0.45\textwidth}
    \centering
    \textbf{Sayan Deb Sarkar} \\
    \normalfont Stanford University
  \end{minipage}
  \hfill
  \begin{minipage}[t]{0.45\textwidth}
    \centering
    \textbf{Sinisa Stekovic} \\
    \normalfont ENPC, IP Paris
  \end{minipage}
  \vspace{1em} \\
  \begin{minipage}[t]{0.45\textwidth}
    \centering
    \textbf{Vincent Lepetit} \\
    ENPC, IP Paris
  \end{minipage}
  \hfill
  \begin{minipage}[t]{0.45\textwidth}
    \centering
    \textbf{Iro Armeni} \\
    Stanford University
  \end{minipage}
}

\begin{document}

\maketitle

\begin{abstract}
Transferring appearance to 3D assets using different representations of the appearance object--such as images or text--has garnered interest due to its wide range of applications in industries like gaming, augmented reality, and digital content creation. However, state-of-the-art methods still fail when the geometry between the input and appearance objects is significantly different. A straightforward approach is to directly apply a 3D generative model, but we show that this ultimately fails to produce appealing results. Instead, we propose a principled approach inspired by universal guidance. Given a pretrained rectified flow model conditioned on image or text, our training-free method interacts with the sampling process by periodically adding guidance. This guidance can be modeled as a differentiable loss function, and we experiment with two different types of guidance including part-aware losses for appearance and self-similarity. Our experiments show that our approach successfully transfers texture and geometric details to the input 3D asset, outperforming baselines both qualitatively and quantitatively. We also show that traditional metrics are not suitable for evaluating the task due to their inability of focusing on local details and comparing dissimilar inputs, in absence of ground truth data. We thus evaluate appearance transfer quality with a GPT-based system objectively ranking outputs, ensuring robust and human-like assessment, as further confirmed by our user study. Beyond showcased scenarios, our method is general and could be extended to different types of diffusion models and guidance functions. Project Page: \textit{\href{https://sayands.github.io/guideflow3d}{https://sayands.github.io/guideflow3d}}
\end{abstract}
\section{Introduction}
\label{sec:intro}
Transferring appearance--including both texture and fine geometric detail--to a 3D object is a challenging and increasingly relevant problem across applications in gaming, augmented reality, and digital content creation. While style transfer has seen substantial progress in the 2D domain \cite{deng2022stytr2,tumanyan2022splicing,tumanyan2023disentangling,wang2023stylediffusion}, extending it to 3D introduces unique challenges due to the irregular, sparse, and diverse nature of 3D representations~\cite{fujiwara2024sn2n, liu2024stylegaussian, vicanerf2023, zhu2025_restyle3d}. Style cues may originate from 3D shapes, 2D images, or natural language, further increasing task complexity. An additional challenge lies in transferring the appearance across objects with substantial geometric differences, required in practical 3D design.

\begin{figure*}[ht!]
    \centering
    \includegraphics[trim=0 0 0 0,clip,width=\linewidth]{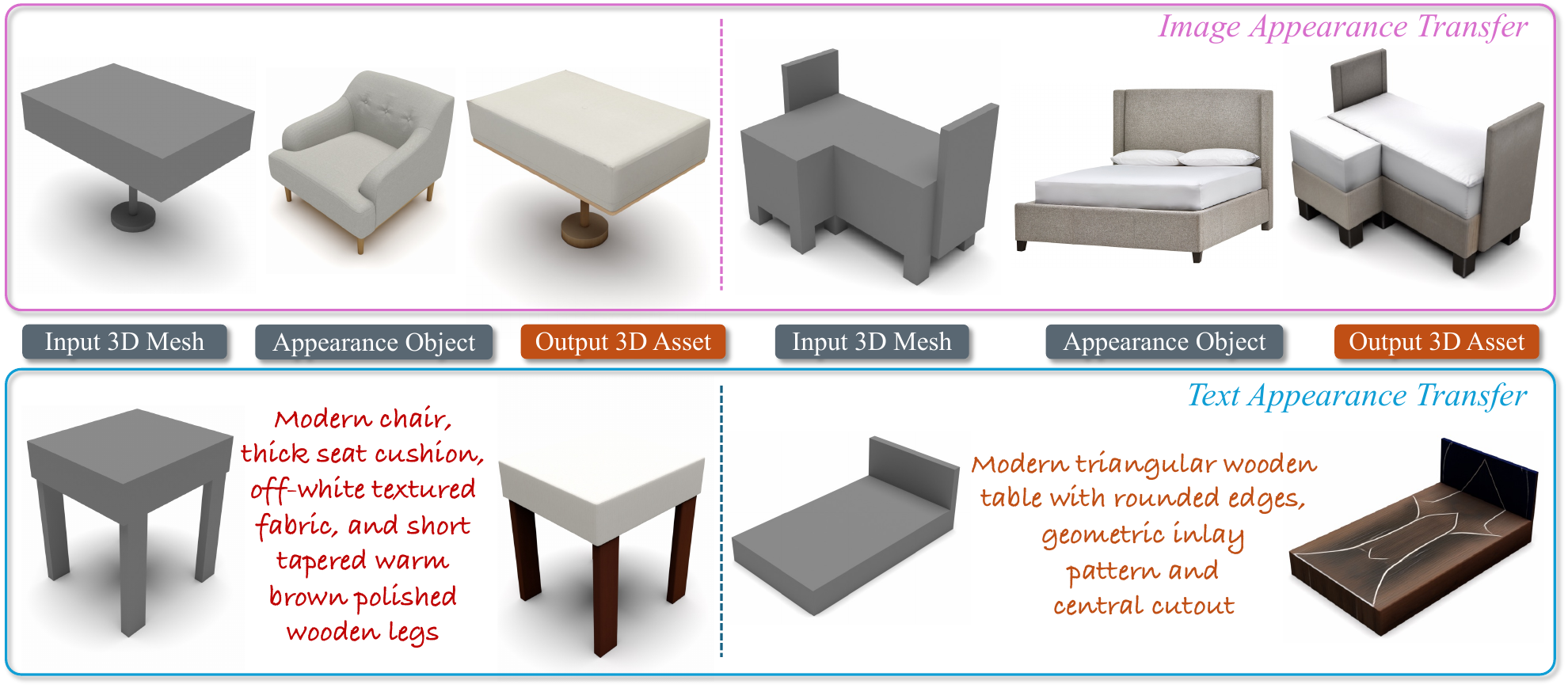}
    \caption{\textbf{\project is a method for 3D appearance transfer robust to strong geometric variations between objects}. Given an input 3D mesh, \eg, designed using simple 3D primitives, it transfers the texture and fine geometric details of an appearance object (\eg, the rounded edges of the table on the top left and the base and mattress distinction of the bed on the top right) but preserves the geometric form of the input mesh. Its flexibility across appearance modalities like meshes or text makes \project efficient for generating diverse 3D assets.}
    \label{fig:teaser}
    \vspace{-15pt}
\end{figure*}

Several existing methods address 3D generation by reformulating it as a multi-view task, leveraging 2D diffusion models and conditioning on, \eg, rendered depth images to preserve input geometry~\cite{oh2023controldreamer, li2024controllable, zhou2023dreampropeller}. However, these methods often produce geometrically inconsistent results due to discrepancies across different views. Recent advances in 3D generative modeling--particularly denoising-based approaches--enable high-quality synthesis of shapes and appearances  conditioned on inputs like text~\cite{poole2022dreamfusion, babuhyperfields,trellis}. Still, these models are constrained by conditioning signals and data distributions used during training. As a result, direct application to appearance transfer yields poor generalization and limited control, particularly when the geometries between the input and appearance objects differ significantly, as we show in our experiments. Note that, in the appearance transfer setting, the input object dictates the global geometry and the appearance one the texture and finer geometric details.

In this work, we present \textit{\project}, a training-free framework for 3D appearance transfer that adaptively steers a pretrained generative model at inference time. Our key insight is that the inductive bias of a pretrained generative model can be repurposed for a new task through \textit{guided rectified flow sampling}, an inference-time mechanism that interleaves flow updates with latent-space optimization. This strategy extends the concept of universal guidance to 3D generation and enables conditioning on objectives for which the base model was not originally trained. Our method builds upon rectified flow models and structured latent representations~\cite{trellis}, and introduces differentiable guidance functions that modulate the generation process without requiring retraining. To address prior work challenges in transferring appearance to an input 3D object while maintaining its global geometry, we propose two novel forms of guidance for appearance transfer to 3D shapes that significantly improve robustness to geometric variations between given input and appearance objects: (i) a part-aware appearance loss, which co-segments the input and appearance 3D shapes into semantically meaningful parts and enforces localized texture and geometry correspondence between the shapes; and (ii) a self-similarity loss, which preserves intrinsic structure within regions of the input during transfer. Our method supports various representations for the appearance object—including mesh-image pairs or text. When a mesh is available, the appearance loss is applied; otherwise, the self-similarity loss guides the generation using either an image or text. In this way, users can control whether appearance affects both geometry and texture (with mesh) or texture alone (with image or text). When only an image is provided, a mesh can be generated via existing methods such as~\cite{trellis}. This flexibility allows \project{} to unify multiple appearance modalities under a single framework, bridging geometric and perceptual style transfer in 3D.

Quantitatively evaluating 3D appearance transfer remains challenging due to the absence of ground truth input shapes that have the transferred appearance and the difficulty of comparing across dissimilar geometries. While metrics like  DINOv2~\cite{oquab2023dinov2}, CLIP~\cite{hessel2021clipscore}, and DreamSim~\cite{fu2023dreamsim} are typically used to assess perceptual similarity and others like PSNR~\cite{Jayasumana2023RethinkingFT}, SSIM~\cite{ssim}, LPIPS~\cite{zhang2018perceptual}, and FID~\cite{cohenbar2025tritexlearningtexturesingle}, for reconstruction quality, they require ground truth data that does not exist in our setting. To address this, we use a GPT-based evaluation scheme \cite{cohenbar2025tritexlearningtexturesingle, wu2023gpteval3d} that performs pairwise output ranking in a human-aligned manner. We further validate the consistency of these rankings through a user study, confirming that GPT-based judgments strongly correlate with human evaluation on this task. Our experiments demonstrate that \project consistently outperforms baselines, generating visually appealing results that accurately reflect the intended style, while also capturing fine-grained geometry. Our contributions are as follows:\vspace{-5pt}
\begin{itemize}
    \item We introduce \textit{\project}, a novel framework for 3D appearance transfer that applies universal, differentiable guidance to a pretrained rectified flow model, enforcing the generation process to respect given constraints for which it was not originally trained. \vspace{-1pt}
    \item We propose part-aware and self-similarity loss functions as effective forms of guidance, enabling localized and structure-preserving style transfer across diverse 3D assets. \vspace{-1pt}
    \item Our method is training-free, generalizable to different appearance representations and, it could be, in principle, extended to a variety of 3D generative models.
\end{itemize}

By decoupling style control from the generation process and enabling inference-time conditioning on novel objectives, \project{} opens new directions in controllable 3D generation and asset stylization. We make our code and benchmark publicly available on our \textit{\href{https://sayands.github.io/guideflow3d}{project website}}.
\section{Related Work}
\label{sec:related_work}

\noindent \textbf{3D Generative Models.} A common method for learning 3D shape generation uses autoencoders, where an encoder maps a point cloud to an embedding, which a decoder reconstructs into a 3D shape. Previous approaches rely on normalized point flows~\cite{yang2019pointflow} or gradient fields~\cite{cai2020learning}. GAN formulations are common in 3D shape generation~\cite{wu2016learning,pavllo2021learning,chan2022efficient,gao2022get3d}. Luo et al. \cite{luo2021diffusion} and more recent approaches~\cite{cheng2023sdfusion,zheng2023lasdiffusion,chen2023single,muller2023diffrf} formulate 3D asset generation as a probabilistic diffusion processes. Some works take advantage of 2D generative models to create 3D assets~\cite{poole2022dreamfusion,tang2023dreamgaussian,lin2023magic3d}. Others \cite{chen2024meshanything,wang2024llama,siddiqui2024meshgpt} propose auto-regressive models for generating 3D meshes. As we discuss further in Sec.~\ref{sec:preliminaries}, Trellis~\cite{trellis} uses rectified flow formulation to learn the generation of structured 3D latents that can effectively be decoded in different 3D representations. Due to its ability to capture fine details in both geometry and texture, we develop our appearance transfer framework around this model.   

\noindent \textbf{2D Style Transfer.} Transferring style between images has been extensively studied in computer vision. Early work~\cite{gatys2015neural} formulates it as an optimization problem over pretrained CNN features, later replaced by efficient feed-forward networks~\cite{johnson2016perceptual,ulyanov2016texture}. With multimodal embeddings, StyleCLIP~\cite{patashnik2021styleclip} enables text-driven style manipulation. Diffusion-based approaches~\cite{zhang2023inversion,wang2023stylediffusion,yang2023paint,qi2024deadiff} now dominate the field, offering fine-grained control through score-based modeling. Cross-Image Attention~\cite{crossimageattention} performs zero-shot appearance transfer by injecting cross-image attention into the diffusion process, while MambaST~\cite{mambast} leverages state-space models for efficient and expressive structure–style fusion. ~\cite{tumanyan2022splicing} trains a generator on a specific content–style example, where structure and appearance constraints are derived from pretrained vision transformer features. Such instance-specific formulations highlight the challenge of achieving generalizable appearance transfer across diverse geometries -- a limitation our approach directly addresses by introducing a guided generative mechanism that
preserves geometric consistency while adapting fine-grained style cues.

\noindent \textbf{3D Style Transfer.} Style transfer has been explored across diverse 3D representations, including point clouds~\cite{yin20213dstylenet}, meshes~\cite{bokhovkin2023mesh2tex}, implicit fields~\cite{huang2021learning,chiang2022stylizing,fan2022unified,liu2023stylerf}, and Gaussian splatting~\cite{liu2024stylegaussian}. 3DStyleNet~\cite{yin20213dstylenet} separates geometry and texture through dual networks, while Mesh2Tex~\cite{bokhovkin2023mesh2tex} maps style images into a learned texture manifold. Recent methods such as StyleRF~\cite{liu2023stylerf} and StyleGaussian~\cite{liu2024stylegaussian} represent major progress but operate on implicit or point-based representations, producing render-only outputs that lack explicit mesh control and often require multi-view optimization. StyleGaussian~\cite{liu2024stylegaussian} stylizes only color while keeping geometry fixed, with performance degrading as Gaussian count increases. In contrast, \project{} performs training-free stylization in a structured latent space, enabling direct mesh editing, part-aware control, and constant inference cost. In diffusion-based 3D stylization, ControlNet-style adapters~\cite{controlnet} have become standard for injecting additional cues such as images, edges, depth, or pose into pretrained models. For example, TEXTure~\cite{richardson2023texture} employs score distillation for text-guided texturing while EASI-Tex~\cite{easitex} extends this with depth-aware conditional diffusion. TriTex~\cite{cohenbar2025tritexlearningtexturesingle} learns a semantic texture field for single-mesh appearance transfer. While effective, these methods are tied to specific training setups and conditioning modalities, limiting their generality. Building on the broader line of diffusion guidance research~\cite{dhariwal2021diffusion,Bansal2023UniversalGF,yu2023freedom}, \project{} combines the concept of universal guidance and rectified flow model from Trellis~\cite{trellis}. By incorporating geometric priors from PartField~\cite{partfield2025}, our method achieves robust and geometry-aware appearance transfer without the need for task-specific finetuning.

\section{Preliminaries}
\label{sec:preliminaries}
\vspace{-10pt}
\noindent \textbf{Structured Latent From 3D Assets.} The geometry and appearance of a 3D object mesh $\mathcal{O}$ can be encoded using a structured latent (\textsc{SLat}) representation $\mathbf{z}$~\cite{trellis}, composed of local latent codes anchored on a 3D voxel grid. This representation is defined as:
\begin{equation}
    \label{eq:slat}
    \mathbf{z} = \{(z_i, p_i)\}_{i=1}^L, \quad z_i \in \mathbb{R}^C, \quad p_i \in \{0, 1, \ldots, N - 1\}^3 \> ,
\end{equation}
where $p_i$ denotes the position of an active voxel intersecting the surface of $\mathcal{O}$, and $z_i$ is the latent vector associated with that voxel. $N$ is the spatial resolution of the grid and $L$ is the number of active voxels, intersecting with the object's surface. 
Since the number of active voxels is significantly lesser than that of a full grid, this representation is computationally less demanding, allowing to work at a higher resolution. The set of active voxel positions $p_i$ outlines the coarse structure of the object, while the corresponding latents $z_i$ capture fine-grained geometric and visual features, as shown in~\cite{trellis}. This  representation thus effectively captures both the global shape and detailed surface characteristics of the object. In practice, following Trellis~\cite{trellis}, multi-view images of $\mathcal{O}$ are rendered, and DinoV2~\cite{oquab2023dinov2,darcet2023vitneedreg} features are extracted for each image. After back-projection and aggregation of these features per voxel, a shallow transformer encoder~\cite{ldm,li2023generalized} compresses them into \textsc{SLat}. Structured latents can be decoded into 3D Gaussians~\cite{kerbl3Dgaussians}, Radiance Fields~\cite{mildenhall2021nerf}, or Meshes~\cite{shen2023flexicubes} using decoders $\mathcal{D}_{3D}$ that share a common architecture, differing only in output layers and trained with representation-specific losses. Please note that, in our case, voxel positions $p_i$ remain fixed to preserve the coarse input geometry, while only the latent codes $z_i$ are steered during generation.

\noindent \textbf{Universal Diffusion Guidance.} We briefly introduce the concept of universal guidance from~\cite{Bansal2023UniversalGF}. Diffusion models are generative models that reverse the process of adding noise, typically Gaussian noise, to the data over time steps. A noisy version of the input $x_0$ at time step $t$ is defined as:
\begin{equation}
    x_t = \sqrt{\beta_t} \, x_0 + \sqrt{1 - \beta_t} \mathcal{N}(0, \mathbf{I}) \> ,
\end{equation}
where $\beta_t$ represents the noise scale at time step $t$,  determined by a scheduling mechanism. Hence, $x_t$ is a weighted sum of the original image and Gaussian noise. As $t$ increases, $\beta_t$ decreases, and $x_t$ approaches pure noise as $t \to T$. With some generalization, a trained diffusion network $f_\theta(x_t)$ reverses the process by predicting the added noise. The denoising process can then be defined as:
\begin{equation}
    \hat{x}_{t - 1} = \hat{x}_t + f_\theta(\hat{x}_t) \>
\end{equation}

While diffusion models can be typically guided by adding a condition to the diffusion model $f_\theta(x_t, \mathbf{c})$, \eg, $\mathbf{c}$ can be text or an image, this limits applications to more general conditions. In universal guidance from~\cite{Bansal2023UniversalGF}, any differentiable guidance function $\mathcal{L}(x_t, \cdot)$ can be used to condition diffusion. Then, the update in the reverse step is defined as:
\begin{equation}
    \hat{x}_{t - 1} = \hat{x}_t + f_\theta(\hat{x}_t, \mathbf{c}) + \nabla_{\hat{x}_t} \mathcal{L}(\hat{x}_t, \cdot) \>
\end{equation}
Such an approach enables general conditioning without altering the diffusion training process.

\begin{figure*}[t]
    \centering
    \includegraphics[trim=0 0 0 0,clip,width=\linewidth]{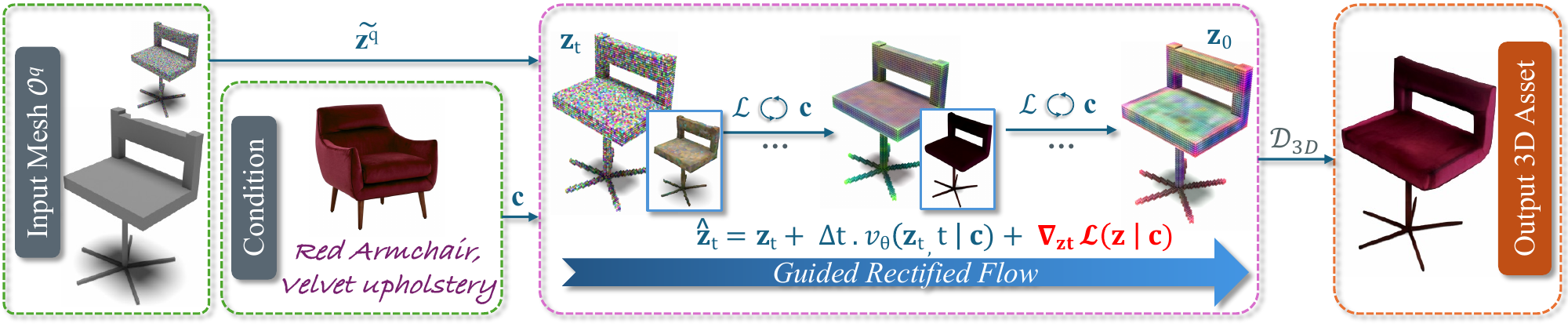}
    \caption{\textbf{\project introduces guided rectified flow for appearance transfer between input object $\mathcal{O}^q$ and an appearance object.} We extend the denoising process of structured latents $\tilde{z}^q$, conditioned by $\mathbf{c}$, by introducing an objective function $\mathcal{L}$ that enforces strong geometric and semantic priors during the process. We show denoised structured latents at different stages of the process, along with corresponding meshes decoded using a pretrained decoder $\mathcal{D}_{\textit{3D}}$. The output 3D asset displays robustness to strong geometric variations between input and appearance objects.}
   \label{fig:architecture_overview}
   \vspace{-10pt}
\end{figure*}

\section{\project}
\label{sec:method}

We present an overview of our approach for appearance transfer in Fig.~\ref{fig:architecture_overview}. Given an input 3D object mesh $\mathcal{O}^q$ and an appearance object $\mathcal{O}^a$, we would like to modify the appearance of $\mathcal{O}^q$ based on $\mathcal{O}^a$, while respecting the geometric structure of $\mathcal{O}^q$. $\mathcal{O}^a$ can be represented as an image-mesh pair or text. We assume here that the appearance object provides the mesh alongside the image. In practice, the mesh can be generated by running \cite{trellis} on the image. We first discuss how we use structured latents from~\cite{trellis} for our appearance transfer problem. Then, we show how we can guide the rectified flow sampling using our objectives to transfer appearance while accounting for structural consistency.

\begin{figure*}[t]
    \centering
    \includegraphics[trim=0 0 0 0,clip,width=\linewidth]{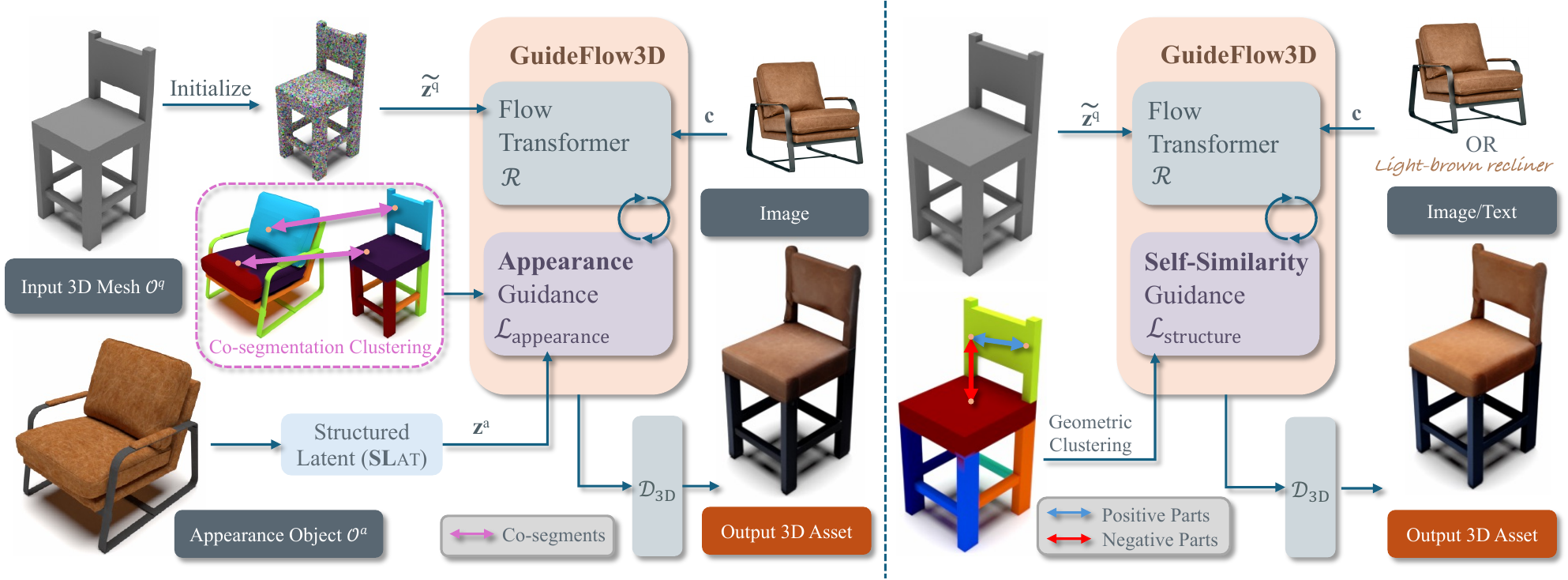}
    \begin{tabular}{cc}
         (a) & \hspace{6cm} (b) \\
    \end{tabular}
    \caption{\textbf{\project with different guidance objectives.} (a) When a textured mesh is available for appearance object $\mathcal{O}^a$, we use our co-segmentation based objective $\mathcal{L}_{appearance}$ to guide appearance transfer. It encourages consistency between structured latents $z^a$ and noisy latents $\tilde{z}^q$. In such case, we use an image of object $\mathcal{O}^a$ to condition the generative model $\mathcal{R}$. (b) When textured mesh is not available, we use our geometric clustering based objective $\mathcal{L}_{structure}$ for guidance. It encourages intra-cluster similarity and inter-cluster disparity when denoising $\tilde{z}^q$. We use text or image to condition $\mathcal{R}$ in such case. For both cases, we use decoder $\mathcal{D_{\textit{3D}}}$ to obtain the output 3D asset.} 
    \label{fig:architecture}
    \vspace{-10pt}
\end{figure*}

\subsection{Structured Latents for Appearance Transfer}
We leverage the latent structure from~\cite{trellis} to represent object $\mathcal{O}^q$, but in general, different representations could be considered. We initialize structured latents $\mathbf{\tilde{z}^q} = \{(\tilde{z}^q_i, p^q_i)\}_{i=1}^{L_q}$ for the input object $\mathcal{O}^q$, where $p^q_i$ are the positional indices of the active voxels, as in Eq.~\eqref{eq:slat}. The $\tilde{z}^q_i$'s are initialized by sampling from a normal distribution. In practice, $p^q_i$ is not a trainable parameter but the $\tilde{z}^q_i$'s are learnable in our appearance transfer formulation. This effectively enforces global geometric consistency of the output with $\mathcal{O}^q$. Following, we define two appearance transfer objectives (Fig.~\ref{fig:architecture}).

\textbf{Appearance-based objective.} When the appearance object is an image-mesh pair, we can extract latents $\mathbf{z^a}$ for the mesh of $\mathcal{O}^a$ using a sparse VAE encoder~\cite{trellis}. Ideally, we seek an `oracle' mapping that aligns each input latent $\tilde{z}^q_i$ with a semantically and geometrically corresponding appearance latent in $\mathbf{z^a}$: \eg, in Fig.~\ref{fig:architecture}~(a),  we would like the $\tilde{z}^q_i$ for voxels of the back leg of an input chair to be mapped to the $z^a_j$ for the voxels of the back leg of the appearance chair. Since such correspondences are not readily available, we approximate them by assigning each $\tilde{z}^q_i$ to its nearest neighbor in $\mathbf{z}^a$ based on feature similarity. Formally, our objective for appearance transfer is defined as:
\begin{equation}
\label{eq:appear}
    \mathcal{L}_\text{appearance} = \frac{1}{L_q} \sum_{i=1}^{L_q} \| \tilde{z}^q_i - z^a_m \|_2^2 \> ,
\end{equation}

where $m$ is the index of the corresponding feature in structured latents $\mathbf{z^a}$ for $\tilde{z}^q_i$. While it could be possible to  extract correspondences based on these latents, they are not trained to be part-aware, and this would lead to false assignments, as shown in Sec.~\ref{sec:ablation_study}. Instead, we establish correspondences based on geometric co-segmentation clustering using part-based feature fields from~\cite{partfield2025}. In practice, we compute PartField~\cite{partfield2025} features per voxel and run $k$-means clustering, thus, relying on approximate matching rather than one-to-one correspondences.

\textbf{Self-similarity objective.} In the second scenario, we assume that the mesh is not available, hence $\mathcal{O}^a$ is represented as an image or text. We observe that structure-based self-similarity descriptors have been shown to effectively capture structural information while being invariant to appearance~\cite{shechtman2007localselfsim, amir2021deep, kolkin2019style}. We, thus, rely on a loss that encourages similarity between same object parts while promoting inter-part separability in the feature space. As the appearance mesh is not available, this loss aligns structural features with the part-aware semantics implied by the textual description or image for the source. To do so, we first perform geometric clustering that assigns each voxel $p_i^q$ to a cluster $\mathcal{C}_q(i)$. Given $\mathcal{O}^q$, for voxels $p_i$ and $p_j$ $\in \mathbf{z_q}$, we compute pairwise cosine similarity between geometric features $\text{sim}_{ij}$ and define a part-aware contrastive loss based on self-similarity. Specifically, for each voxel $p_i$, the set of positive samples is defined as all $j \neq i$ such that $\mathcal{C}_q(i) = \mathcal{C}_q(j)$, and all other voxels serve as negatives. The objective can then be defined by:
\begin{equation}
\label{eq:structure}
\mathcal{L}_\text{structure} = - \frac{1}{L_q} \sum_{i=1}^{L_q} \log \frac{\sum\limits_{ j \in \mathcal{C}_q(i), j \neq i,} \exp(\text{sim}_{ij})}{\sum\limits_{j \in \mathcal{C'}_q(i)} \exp(\text{sim}_{ij})} \>
\end{equation} 
where $L_q$ denotes the set of all voxels and $\mathcal{C'}$ is the complement set of $\mathcal{C}$. Due to its inherent contrastive nature, $\mathcal{L}_\text{structure}$ promotes local consistency without homogenizing appearance globally.

\subsection{Guiding Structured Latents for Appearance Transfer}
Our objective is defined using a Bayesian formulation. Given the input 3D object mesh $\mathcal{O}^q$ and appearance object $\mathcal{O}^a$, we would like to maximize the posterior $P(\mathcal{O}^q | \mathcal{O}^a)$:

\begin{equation}
    \log P(\mathcal{O}^q | \mathcal{O}^a) = \log P(\mathcal{O}^q) + \log P(\mathcal{O}^a | \mathcal{O}^q) \> ,
\end{equation}

\begin{wrapfigure}{r}{0.52\textwidth}
  \centering
  \vspace{-12pt}
  \includegraphics[width=0.45\textwidth]{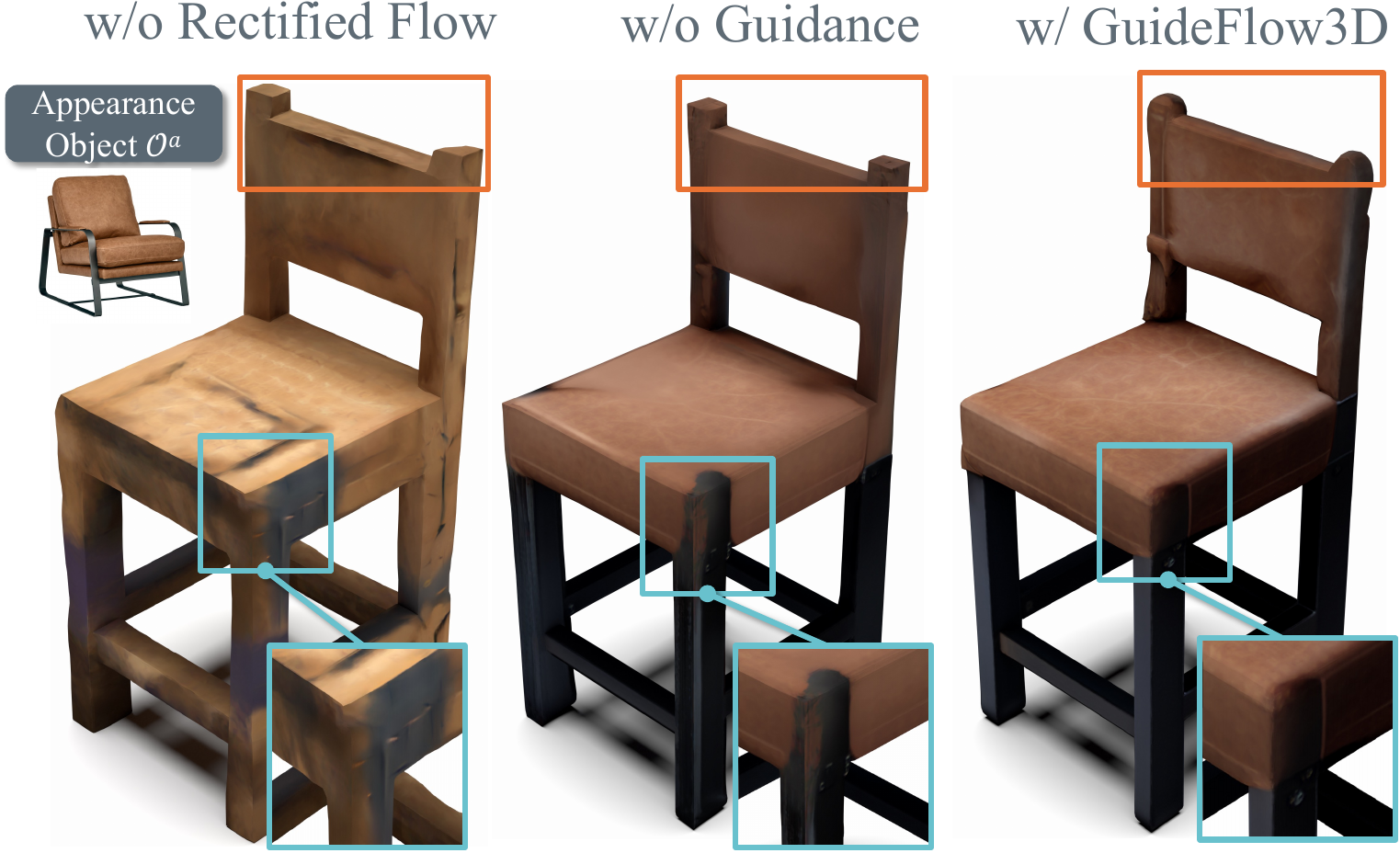}
  \begin{tabular}{ccc}
    \hspace{0.5cm} (a) & \hspace{1cm} (b) & \hspace{1cm} (c) \\
 \end{tabular}
  \caption{\textbf{Effect of different modules.} (a) Using only optimization with our objective functions to transfer appearance is insufficient as it does not enforce realistic distribution over the structured latent space. (b) The rectified flow model from $\cite{trellis}$ fails to transfer appearance when appearance and input objects have significantly different geometries. (c) We obtain appealing 3D assets when using rectified flow guidance of our \project.}
   \label{fig:guidance_explain}
   \vspace{-10pt}
\end{wrapfigure}

where prior $\log P(\mathcal{O}^q)$ models the geometric prior of $\mathcal{O}^q$ while  likelihood $\log P(\mathcal{O}^a | \mathcal{O}^q)$ models the appearance. The objectives defined in the previous section are not enough to perform appearance transfer. While $\mathcal{L}_\text{appearance}$ models the posterior $\log P(\mathcal{O}^q | \mathcal{O}^a)$, solely optimizing it proves insufficient. It tends to shift the latent distribution away from the one modeled by the generative network, often leading to implausible or incoherent results across diverse shapes (Fig.~\ref{fig:guidance_explain}~(a)). Meanwhile, $\mathcal{L}_\text{structure}$ only models $P(\mathcal{O}^q)$ of our Bayesian formulation. To address these issues and better align with the generative prior, we introduce appearance transfer as guidance in rectified flow.

\noindent \textbf{Rectified Flow Guidance.} Rectified flow models~\cite{liu2023flow, albergo2023building, lipman2023flow} define a linear forward process, $\boldsymbol{\mathbf{z}}(t) = (1 - t)\boldsymbol{\mathbf{z}}_0 + t\boldsymbol{\epsilon}$, which interpolates between data samples $\boldsymbol{\mathbf{z}}_0$ and noise $\boldsymbol{\epsilon}$ over time $t$. The corresponding backward process is modeled as a time-dependent vector field $\boldsymbol{v}(\boldsymbol{\mathbf{z}}, t) = \nabla_t \boldsymbol{\mathbf{z}}$, which transports noisy samples back toward the data distribution. This vector field can be approximated by a neural network $\boldsymbol{v}_\theta$, trained via the Conditional Flow Matching~(CFM) objective~\cite{lipman2023flow}:
\begin{equation}
    \mathcal{L}_\text{CFM}(\theta)=\mathbb{E}_{t,\boldsymbol{\mathbf{z}}_0,\boldsymbol{\epsilon}}\|\boldsymbol{v}_\theta(\boldsymbol{\mathbf{z}}, t)-(\boldsymbol{\epsilon}-\boldsymbol{\mathbf{z}}_0)\|^2_2 \> . \label{eq:cfm}
\end{equation}

While rectified flow enables sample generation that remains close to the learned data distribution, using it alone is insufficient for semantic control, particularly when transferring fine-grained appearance or structure. As shown in Fig.~\ref{fig:guidance_explain}~(b), applying rectified flow without additional constraints often yields outputs that are within the learned distribution but fail to reflect the style texture or semantic structure, especially across diverse object categories and shapes. To enforce semantic control while preserving realism, we interleave latent-space optimization with sampling steps from a rectified flow model $\mathbf{v}_\theta(\mathbf{z}, t \mid \mathbf{c})$, conditioned on a global signal $\mathbf{c}$. Based on~\cite{trellis}, the signal $\mathbf{c}$ can either be an image or text. Starting from an initial latent $\mathbf{z}_T \sim \mathcal{N}(0, I)$, reverse-time flow is defined as: \vspace{-10pt}

\begin{equation}
    \hat{\mathbf{z}}_{t} = \mathbf{z}_t + \Delta t \cdot \mathbf{v}_\theta(\mathbf{z}_t, t \mid \mathbf{c}) \>
\end{equation}

where $t \in [0, 1]$ is linearly spaced. We adapt the equation to include guidance: \vspace{-10pt}

\begin{equation}
    \hat{\mathbf{z}}_t = \mathbf{z}_t + \Delta t \cdot \mathbf{v}_\theta(\mathbf{z}_t, t \mid \mathbf{c}) \> + \nabla_{\mathbf{z} t} \mathcal{L}(\mathbf{z} \mid \mathbf{c})
\end{equation}

where $\mathcal{L}$ is a general guidance objective. The optimized latent $\hat{\mathbf{z}}$ is then used for the next flow step. Such optimization-flow scheme allows for flexible conditioning via differentiable objectives without retraining the model, extending universal guidance~\cite{Bansal2023UniversalGF} to any rectified flow model $\mathcal{R}$. The final output can be decoded into the 3D representation of choice based on 3D decoders from~\cite{trellis}. The guidance can be fit into our Bayesian formulation, where conditioned rectified flow models both prior $P(\mathcal{O})$ and likelihood $P(\mathbf{c} | \mathcal{O})$, and the guidance terms become additional factors in the prior and likelihood terms. We find that this interleaved scheme preserves global realism while enabling fine-grained control over both appearance and structure, as shown in Fig.~\ref{fig:guidance_explain}~(c). Qualitative ``in-the-wild’’ transfers (Fig.~\ref{fig:quals_in_the_wild}) further illustrate robustness under large geometric discrepancies, \eg, animal$\rightarrow$furniture.

\noindent \textbf{Condition-Specific Guidance.} Our method supports flexible conditioning depending on the form of the input. When a reference image is provided, we may employ either appearance- or structure-based guidance. For appearance transfer, we assume access to the appearance mesh $\mathcal{O}^a$ from which structured latents $\mathbf{z}^a$ are extracted, and we optimize $\mathcal{L}_\text{appearance}$ to transfer appearance. Alternatively, even in the absence of a mesh, the same reference image can be used to compute geometric features, enabling structure-aware transfer via $\mathcal{L}_\text{structure}$. When the condition is a text prompt, we rely exclusively on $\mathcal{L}_\text{structure}$, using semantic clustering of geometric features to induce part-aware correspondences. This setup enables our framework to seamlessly handle both visual and textual inputs while maintaining geometric fidelity and semantic consistency.
\section{Experiments}
\label{sec:experiment}
\noindent \textbf{Dataset.} Since there are no publicly available datasets for our task of transferring appearance across different shapes, we create a benchmark for evaluation. First, for input mesh, we generate synthetic objects using procedural models from~\cite{stekovic2025pytorchgeonodes}. Second, for the appearance mesh and images, we leverage the ABO dataset~\cite{collins2022abo}, with the text captions provided by~\cite{trellis}. ABO contains $\sim$8K artist-designed 3D models from Amazon, featuring complex geometries and high-resolution materials across 63 categories, mainly focused on furniture and interior decor. We use $5$ categories: \texttt{bed, cabinet, chair, table}, and \texttt{sofa}, randomly sampling $100$ objects from each dataset. Here onwards, we refer to  the dataset of procedurally generated objects as \textit{simple} and the ABO subset as \textit{complex}. Using \textit{simple} and \textit{complex} meshes for the different categories, we create $250$ input-appearance object pairs for each of our $4$ experimental setups: (i) \textit{simple}-\textit{complex} \textit{intra-category}, (ii) \textit{simple}-\textit{complex} \textit{inter-category}, (iii) \textit{complex}-\textit{complex} \textit{intra-category}, and, (iv) \textit{complex}-\textit{complex} \textit{inter-category}.

\noindent \textbf{Evaluation Metrics.} In absence of ground truth transferred texture for the input 3D mesh, encoder-based metrics are not representative of appearance transfer for 3D shapes of very different geometries (see more in Appendix Sec.~\ref{sec:perc_sim_eval}). LLMs have been previously shown to be a human aligned evaluator for generation tasks capturing structural preservation and content alignment~\cite{wu2023gpteval3d, Zhang2023GPT4VisionAA,peng2024dreambench}. We use GPT-5 to carefully develop a more nuanced human-aligned ranking system. This system evaluates results across six criteria--Style Fidelity, Structure Clarity, Style Integration, Detail Quality, Shape Adaptation, and Overall Quality--capturing both semantic intent and the perceptual quality of texture transfer in 3D. Following prior work~\cite{cohenbar2025tritexlearningtexturesingle, wu2023gpteval3d}, we use multi-view renders of input, appearance, and output for all methods and prompt GPT-5 to give us a ranking (details in Appendix Sec.~\ref{sec:gpt_eval}). We show that GPT-5 is aligned with human preferences on this task via a user study in Appendix Sec.~\ref{sec:user_study}.

\noindent \textbf{Baselines.} We compare our \project{} for appearance transfer against multiple baselines: \textbf{(1) UV Nearest Neighbor}: We find the nearest neighbors in Euclidean space for each point in the input mesh $\mathcal{O}^q$ to the mesh of the appearance object $\mathcal{O}^a$, and map the coordinates of the UV texture map accordingly; \textbf{(2) Image-to-3D}: We render the input mesh from multiple views, apply state-of-the-art 2D style transfer models~\cite{crossimageattention,mambast}, and lift the stylized renderings to 3D via the image-conditioned Trellis model; \textbf{(3) EasiTex}~\cite{easitex}: is a conditional diffusion-based method that uses edge-aware conditioning and ControlNet~\cite{controlnet} to texture an existing 3D mesh from a single RGB image, \textbf{(4) Trellis}~\cite{trellis}: serves as our baseline without guidance, enabling detailed local texture transfer using structured latents; and \textbf{(5) Text-to-3D}: For text-based appearance, we first generate a reference image using Stable Diffusion~\cite{sdxl} and then apply the same Image-to-3D pipeline described above.

\subsection{Appearance Transfer}

\begin{figure*}[h!]
    \centering
    \includegraphics[trim=0 0 0 0,clip,width=\linewidth]{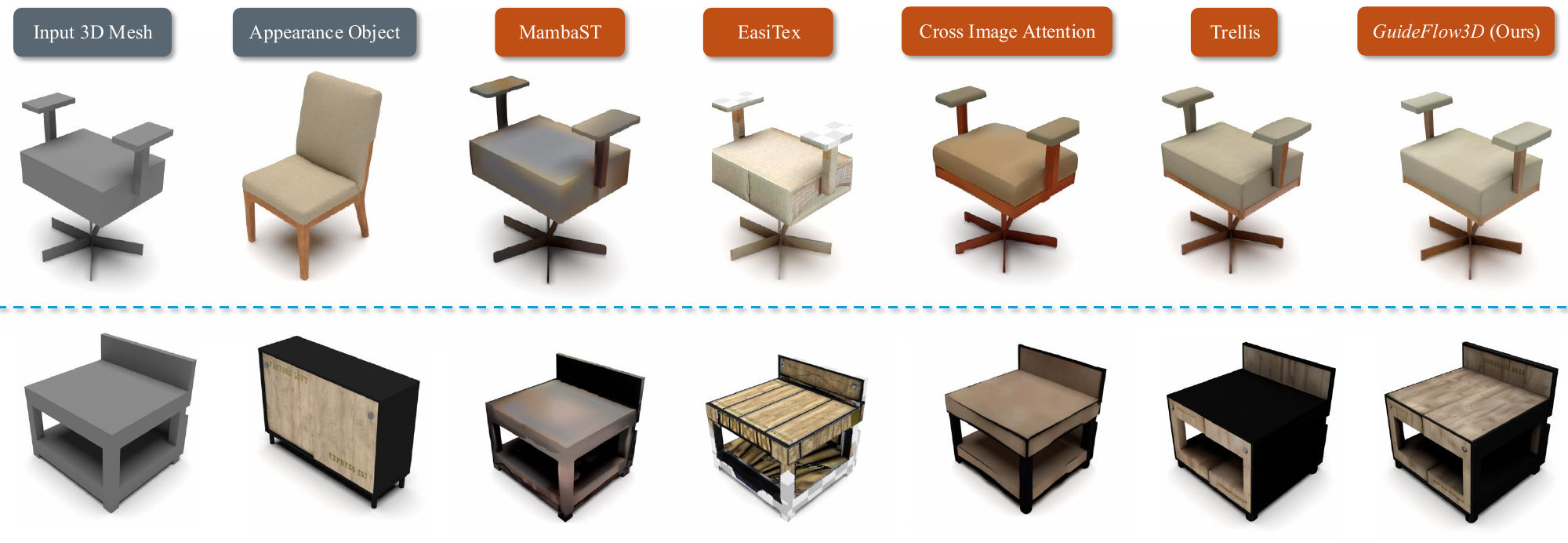}
    \vspace{-15pt}
    \caption{\textbf{Qualitative Comparisons showing quality of appearance transfer.} Top and bottom rows show intra-class (\texttt{chair} to \texttt{chair}) and inter-class (\texttt{cabinet} to \texttt{bunk bed}) results respectively. In both examples, MambaST~\cite{mambast} blends the textures from both input and appearance objects, giving a grey hue to the final result. EasiTex~\cite{easitex} generates non-smooth repetitions of texture and fails to generate textures for the entire object (e.g., the handles of the chair or the bottom part of the bunk bed). Cross Image Attention~\cite{crossimageattention} performs better but omits texture details (cabinet's wood texture) and fails to preserve good local geometry. Trellis~\cite{trellis} preserves better the texture of the appearance object and does better on matching it to the input object, however, it fails at providing a uniform texture on the arms of the chair and does not preserve the overall geometry of the bunk bed by closing the side hole. Ours performs the best by adhering to the appearance object's texture and matching it on the input object, while preserving the overall geometry of the input object.} 
    \label{fig:quals_main}
    \vspace{-5pt}
\end{figure*}

Our first goal is to transfer appearance from an object given as image and textured mesh, or as text, to an untextured 3D object, within the \textit{same} semantic category. This setting allows us to systematically assess how well each method preserves stylistic intent while adapting to varied geometries within each semantic category. Tab.~\ref{tab:main} shows the results on the \textit{simple}-\textit{complex} \textit{intra-category} set under both image-mesh ($\mathcal{L}_\text{appearance}$) and text conditioning ($\mathcal{L}_\text{structure}$). As illustrated in Fig.~\ref{fig:quals_main} (top row), transferring the appearance from a \texttt{chair} to another \texttt{chair} reveals clear differences in quality. MambaST~\cite{mambast} produces textures that are globally consistent due to its state-space backbone, ensuring smooth overall color and material coherence. However, residual gray tones from the input mesh persist, causing uneven blending and inconsistent local texture alignment. Cross Image Attention~\cite{crossimageattention} effectively transfers local appearance patterns through image-conditioned attention, yet fails to maintain consistent mapping when applied to 3D surfaces, introducing artifacts during transfer. EASI-Tex~\cite{easitex} performs competitively with its ControlNet-based edge and depth conditioning, yet struggles under large geometric deviations due to limited generalization beyond its training setup. Trellis~\cite{trellis} improves upon this with its probabilistic structured latent generation but struggles to integrate appearance seamlessly across large geometric variations. Although it associates some textures reasonably (e.g., the chair seat transferred between the two), it often results in loss of fine-grained details and patchy surfaces. In contrast, \project{} achieves strong improvements in both appearance transfer and structural clarity preservation in image and text settings. The transferred appearance aligns well with object structure, as seen
in the smooth material transitions and consistent fabric textures on the \texttt{chair}. This demonstrates
the ability of our approach to effectively translate both visual and semantic cues into high-fidelity,
structure-aware textures across diverse 3D objects (more results in Appendix Sec.~\ref{sec:quals_appendix}).

\begin{table}[ht!]
  \centering
  \caption{Quantitative comparison based on our GPT-based ranking metrics that rank quality of appearance transfer based on different criteria. Results are on the \textit{simple}-\textit{complex} \textit{intra-category} set.}
  \vspace{-5pt}
  \resizebox{\linewidth}{!}{
   \begin{tabular}{l|cccccc}
    \toprule
    & \multicolumn{6}{c}{\textbf{Ranking metrics}} \\
      Methods & Fidelity $\downarrow$ & Clarity $\downarrow$ & Integration $\downarrow$ & Quality $\downarrow$
      & Adaptation $\downarrow$ & Overall $\downarrow$ \\  
     \midrule\arrayrulecolor{black} 
    \multicolumn{7}{l}{\cellcolor{colorturquoise}{\textit{w/ Image Condition ($\mathcal{L}_\text{appearance}$)}}} \\
    UV Nearest Neighbor & 4.12 & 3.84 & 4.30 & 4.10 & 4.43 & 4.33 \\
    MambaST~\cite{mambast} & 4.94 & 3.55 & 4.56 & 4.90 & 4.42 & 4.87 \\
    Cross Image Attention~\cite{crossimageattention} & 3.56 & 3.48 & 3.32 & 3.83 & 3.47 & 3.59 \\
    EasiTex~\cite{easitex} & 3.18 & 4.30 & 4.08 & 3.17 & 4.18 & 3.81 \\
    Trellis~\cite{trellis} & \nd{2.51} & \nd{2.58} & \nd{2.53} & \nd{2.69} & \nd{2.61} & \nd{2.62} \\
    \textit{\project{}} (Ours) & \fs 1.89 & \fs 2.41 & \fs 2.07 & \fs 2.23 & \fs 2.28 & \fs 2.12 \\

    \midrule\arrayrulecolor{black} 
    \multicolumn{7}{l}{\cellcolor{colorturquoise}{\textit{w/ Text Condition ($\mathcal{L}_\text{structure}$)}}} \\
        UV Nearest Neighbor & 3.12 & 3.21 & 3.82 & 3.61 & 3.43 & 3.64 \\
        SDXL + Cross Image Attention & 2.88 & 2.52 & 3.25 & 3.38 & 3.29 & 2.98 \\
        Trellis~\cite{trellis} & \nd{2.01} & \nd{1.89} & \nd{2.67} & \nd{2.75} & \nd{2.55} & \nd{2.39} \\
        \textit{\project{}} (Ours) & \fs 1.54 & \fs 1.63 & \fs 2.01 & \fs 2.15 & \fs 2.44 & \fs 1.95 \\
    \bottomrule
    \end{tabular}
    }
\label{tab:main}
\end{table}

Furthermore, to understand cross-category generalization, we evaluate a more challenging setting where the structure and appearance inputs come from different object categories, \eg, transferring the appearance of a \texttt{cabinet} to a \texttt{bed} (Fig.~\ref{fig:quals_main} , bottom row). Results under image conditioning for both \textit{simple}-\textit{complex} and \textit{complex}-\textit{complex} sets are in Tab.~\ref{tab:all_cat}. We show results for text conditioning in the Appendix (Sec.~\ref{sec:text_guidance}). This scenario tests not only fidelity to style but also the ability to adapt and maintain perceptual coherence under stronger geometric variations. \project{} maintains good performance across all metrics despite the added difficulty. In Fig.~\ref{fig:quals_main}, compared to Cross Image Attention~\cite{crossimageattention} and EASI-Tex~\cite{easitex}, which degrade under large shape discrepancies, our method preserves both global coherence and fine structural fidelity. Trellis~\cite{trellis} not only fails to transfer the \texttt{cabinet}’s appearance effectively but also introduces a major geometry change by closing a hole on the side and texturing the altered region. Our robust performance highlights the ability to disentangle style from structure, enabling us to transfer appearance features even when spatial priors do not align. This demonstrates that our method is applicable to a variety of real-world applications.

\begin{table}[ht!]
  \centering
  \caption{Quantitative comparisons ranking of our \project against baselines for different experimental settings. Results are shown with image conditioning ($\mathcal{L}_\text{appearance}$).}
  \vspace{-5pt}
  \resizebox{\linewidth}{!}{
   \begin{tabular}{l|ccc|ccc}
    \toprule
    & \multicolumn{6}{c}{\textbf{Ranking metrics}} \\
      Methods & Fidelity $\downarrow$ & Clarity $\downarrow$ & Adaptation $\downarrow$
      & Fidelity $\downarrow$ & Clarity $\downarrow$ & Adaptation $\downarrow$  \\  
     \midrule\arrayrulecolor{black} 
      & \multicolumn{3}{c}{\cellcolor[HTML]{EEEEEE}{\textit{Intra-Category}}} &
        \multicolumn{3}{c}{\cellcolor[HTML]{EEEEEE}{\textit{Inter-Category}}} \\
        \multicolumn{7}{l}{\cellcolor{colorturquoise}{\textit{Simple-Complex}}} \\
        UV Nearest Neighbor & 4.12 & 3.84 & 4.43 & 4.06 & 3.51 & 4.17 \\
        MambaST~\cite{mambast} & 4.94 & 3.55 & 4.42 & 4.87 & 3.57 & 4.38 \\
        Cross Image Attention~\cite{crossimageattention} & 3.56 & 3.48 & 3.47 & 3.54 & 3.55 & 3.52 \\
        EasiTex~\cite{easitex} & 3.18 & 4.30 & 4.18 & 3.25 & 4.21 & 4.10 \\
        Trellis~\cite{trellis} & \nd{2.51} & \nd{2.58} & \nd{2.61} & \nd{2.64} & \nd{2.85} & \nd{2.76} \\
        \textit{\project{}} (Ours) & \fs 1.89 & \fs 2.41 & \fs 2.28 & \fs 1.99 & \fs 2.75 & \fs 2.45 \\
    \midrule\arrayrulecolor{black} 
    \multicolumn{7}{l}{\cellcolor{colorturquoise}{\textit{Complex-Complex}}} \\
        UV Nearest Neighbor & 3.31 & 3.11 & 3.41 & 3.54 & 2.99 & 3.49 \\
        Cross Image Attention & 4.00 & 4.13 & 4.07 & 3.79 & 3.99 & 3.91 \\
        MambaST & 4.63 & 3.88 & 3.42 & 4.54 & 3.33 & 3.92 \\
        EasiTex & 3.29 & 4.21 & 4.23 & 3.19 & 4.26 & 4.21 \\
        Trellis~\cite{trellis} & \nd{2.82} & \nd{2.73} & \nd{2.81} & \nd{2.99} & \nd{3.15} & \nd{3.09} \\
        \textit{\project{}} (Ours) & \fs 2.21 & \fs 2.31 & \fs 2.34 & \fs 2.24 & \fs 2.69 & \fs 2.56 \\
    \bottomrule
    \end{tabular}
    }
\label{tab:all_cat}
\vspace{-0.6cm}
\end{table}

\subsection{Application: In-the-wild Appearance Transfer}

\begin{figure*}
    \centering
    \includegraphics[trim=0 0 0 0,clip,width=\linewidth]{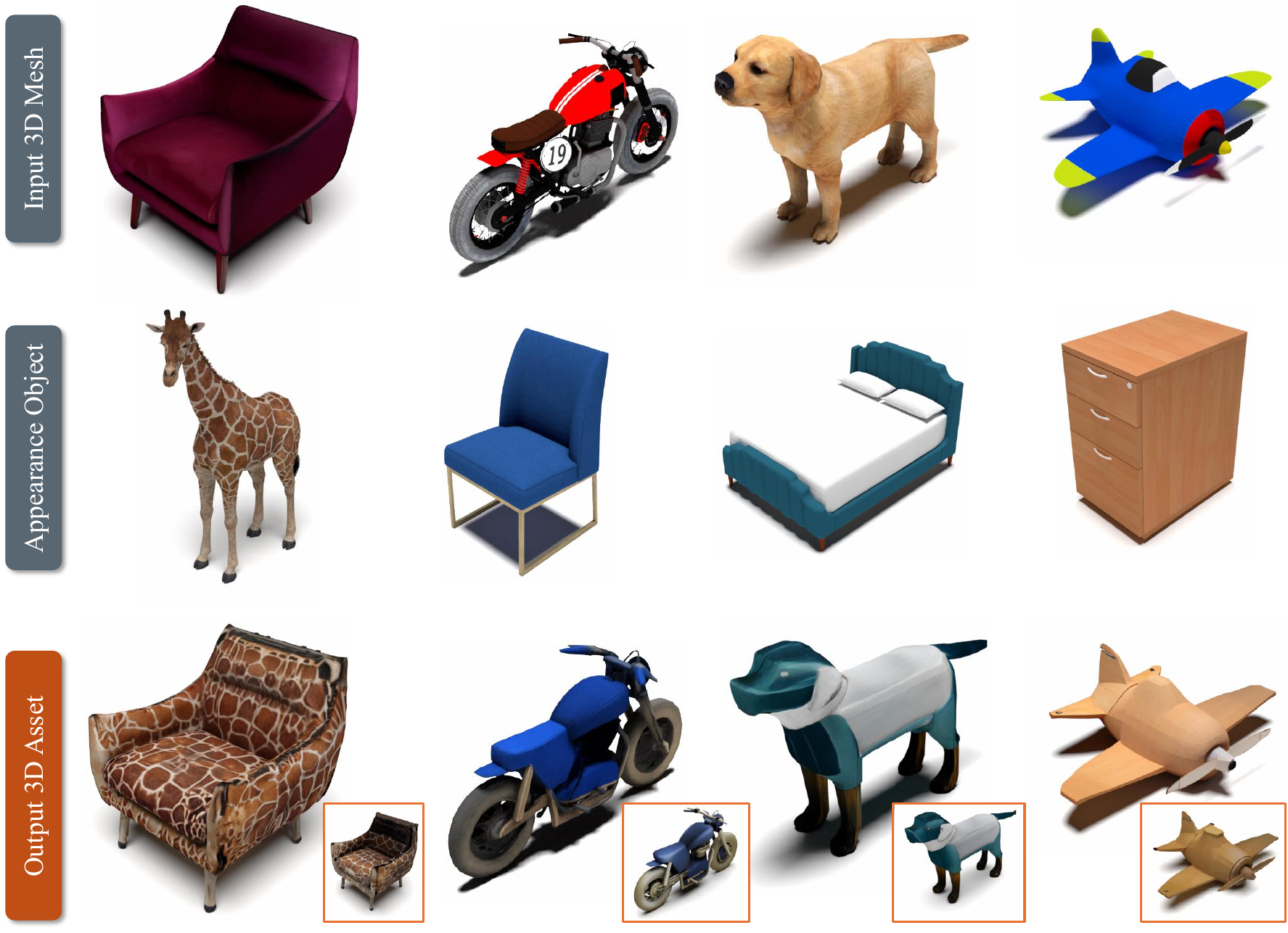}
    \caption{\textbf{Qualitative Comparisons showing in-the-wild appearance transfer.} Our \project robustly transfers appearance between diverse semantic categories. Patterns from the body of the giraffe are transferred to the seat, arms and backrest of the chair, while the appearance of giraffe legs is transferred to chair legs. Across other categories we observe interesting appearance transfers: chair legs to bike wheels, bed legs to golden retriever legs, and cabinet handles to airplane propeller. The smaller insets in \textcolor{orange}{orange} boxes show Trellis~\cite{trellis} results, which exhibit weaker structural grounding and inconsistent material mapping, demonstrating \project{}’s advantage in preserving geometry-aware texture alignment even under large semantic shifts.}
    \label{fig:quals_in_the_wild}
    \vspace{-10pt}
\end{figure*}

To assess the generalization of \project beyond furniture-to-furniture, we explore appearance transfer in-the-wild using structurally diverse 3D assets from Objaverse-XL~\cite{objaverseXL} and ABO~\cite{collins2022abo}, where either dataset can provide the input or appearance object. This setting poses significant robustness challenges due to extreme variation in object categories, mesh quality, and geometric complexity. In this setting, appearance objects are represented as mesh-image pairs, allowing us to use the part-aware appearance loss $\mathcal{L}_\text{appearance}$, as guidance. As illustrated in Fig.~\ref{fig:quals_in_the_wild}, our approach successfully transfers texture and fine geometric details while preserving the original structure. It transfers the appearance between objects in various semantic categories, such as animals, vehicles, and furniture. Despite the lack of semantic alignment or category overlap, we are able to apply texture in a part-aware and structurally consistent manner, highlighting the strength of co-segmentation and geometric clustering-based guidance mechanism. The resulting outputs show high visual fidelity with textures that adhere naturally to the surface topology of the target mesh. Compared to Trellis~\cite{trellis}, which often loses part-aware texture continuity under large semantic shifts, \project{} maintains structurally aligned and visually coherent material transfer across categories. Notably, this application does not require additional assumptions or adaptations, and generalizes to unseen shape categories and object styles, reinforcing the adaptability of \project. The ability to generate visually coherent, stylized outputs has relevance for 3D asset stylization in AR/VR and digital twin generation.

\label{sec:in_the_wild_app_transfer}

\section{Conclusion}
\label{sec:conclusion}

Our \project shows promising results in the field of 3D appearance transfer, possibly enabling exciting new applications. Given that our approach can generate realistic 3D assets from simplistic CAD designs, it could play a role in democratizing and simplifying 3D content creation. Tools for 3D content creation would become more accessible to artists which would significantly reflect on rapid development, prototyping, and creativity in XR and gaming platforms for example. 

\noindent \textbf{Limitations and Future Work.} Our method is optimization based, thus it is not meant for real-time use cases. Our runtime is $96$s on an NVIDIA 4090 GPU, compared to $78$s for baseline~\cite{trellis}. In the future, we could train self-supervised feed-forward models for faster inference. Our implementation depends on the performance of~\cite{trellis} and~\cite{partfield2025}, and failures of these models would impact our performance. Our approach assumes noiseless meshes which is a limiting factor for some future applications scenarios. Developing novel guidance objectives for new applications is an interesting future research direction to address such scenarios. The scope of our main experiments includes furniture objects from ~\cite{collins2022abo} and~\cite{stekovic2025pytorchgeonodes}, but in practice, our method can be applied to a variety of object categories, as shown in Sec.~\ref{sec:in_the_wild_app_transfer}, and downstream tasks such as scene editing (see Appendix Sec.~\ref{sec:scene_editing}).

\noindent \textbf{Ethical Considerations.} Next to the exciting possibilities, there are considerable risks that should be addressed including manipulation and Deepfakes for spreading misinformation, concerns regarding intellectual property, and bias amplifications. Ethical usage of our method includes aspects of disclosing when 3D content is generated using AI, respecting and attributing source content licenses, and building systems for understanding biases are some of the ways for tackling these issues.  
\section{Acknowledgements}
\label{sec:ack}

We thank Nicolas Dufour and Arijit Ghosh from Imagine Labs for helpful discussions on universal guidance, and Liyuan Zhu and Jianhao Zheng from Gradient Spaces Research Group for help with conducting the user study.

{
\small
\bibliography{references}
}

\clearpage
\appendix
\setcounter{section}{0}
\counterwithin{table}{section}
\renewcommand{\thesection}{\Alph{section}}

{\large \textbf{Appendix}\\[0.4cm]}
\textit{In the appendix, we provide the following:
\begin{itemize}[leftmargin=12pt,itemsep=0em]
    \item Intra- and inter-category results using text condition with $\mathcal{L}_\text{structure}$ (Sec.~\ref{sec:text_guidance})
    \item Results on using a rendered image as a condition with $\mathcal{L}_\text{appearance}$ (Sec.~\ref{sec:render_guidance})
    \item Ablation study on design choices (Sec.~\ref{sec:ablation_study})
    \item Application of \project{} on scene editing (Sec.~\ref{sec:scene_editing})
    \item Experimental and evaluation setup details (Sec.~\ref{sec:exp_details})
    \item Issue with perceptual similarity evaluation metrics (Sec.~\ref{sec:perc_sim_eval})
    \item Results of Human Evaluation (Sec.~\ref{sec:user_study})
    \item Details and analysis of the GPT-based evaluation setup (Sec.~\ref{sec:gpt_eval})
    \item Additional qualitative results (Sec.~\ref{sec:quals_appendix})
\end{itemize}
}

\section{Text Conditioning with $\mathcal{L}_\text{structure}$}
\label{sec:text_guidance}
\begin{figure*}
    \centering
    \includegraphics[trim=0 0 0 0,clip,width=\linewidth]{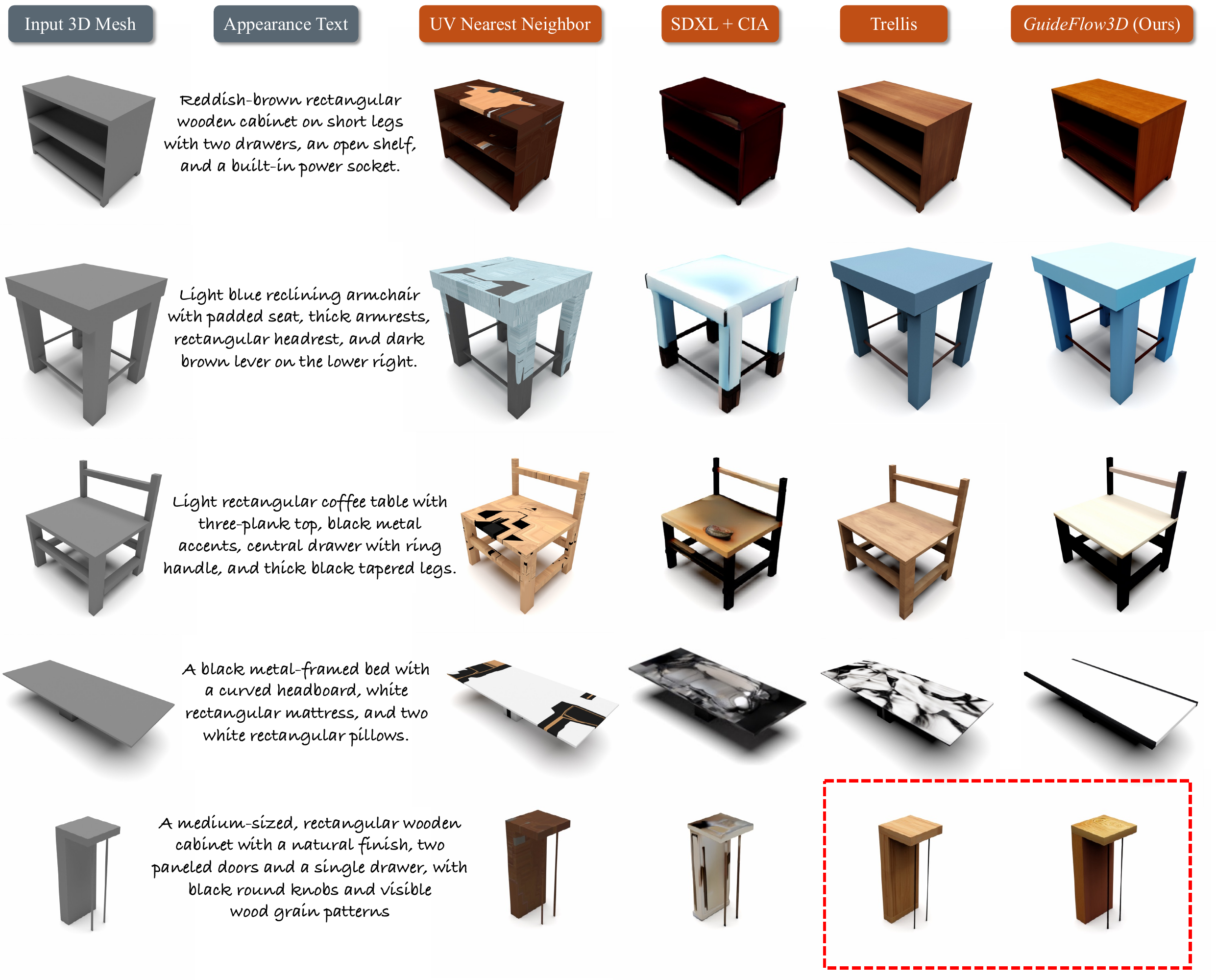}
    \caption{\textbf{Qualitative Comparisons showing results using text condition with $\mathcal{L}_\text{structure}$.} Each row shows an input 3D mesh, a descriptive appearance text, and textured results generated by different methods. We show challenging examples with considerable discrepancies between the appearance text prompts and input geometries. \project{} grounds textual descriptions in geometry, producing coherent, part-aware textures. The fifth row illustrates a failure case where abstract terms like “metal” or “cushioned” are not fully captured, highlighting the difficulty of interpreting underspecified text. Overall, \project{} delivers detailed, realistic appearances aligned with object semantics.}
    \label{fig:suppl_quals_text}
\end{figure*}

Similar to image conditioned results in Tab.~\ref{tab:all_cat}, we report results on using text prompts as the condition $\mathbf{c}$ of the rectified flow model in Tab.~\ref{tab:all_cat_text}. As shown in Fig.~\ref{fig:suppl_quals_text}, UV Nearest Neighbor fails to capture the described materials or object semantics, often ignoring features such as drawers, metal frames, or color attributes. SDXL + Cross-Image Attention~\cite{crossimageattention} produces locally coherent patterns but lacks structural grounding, resulting in oversaturated or unrealistic textures. Trellis~\cite{trellis} improves the plausibility of the output yet misses fine-grained attributes—such as metallic elements, accent colors, or multi-material compositions and frequently distorts local part boundaries. In contrast, \project{} achieves consistently superior performance across all ranking metrics, both in \textit{intra-category} and \textit{inter-category}. In particular, our method preserves structural clarity and convincingly adapts texture, even when appearance is only described in abstract terms. The intra-category results reveal \project's capacity to maintain semantic coherence between structure and style despite geometric variation. More strikingly, in the inter-category setting, where structure and appearance belong to different object types, our method still maintains better ranking scores (i.e., higher quality), demonstrating its robustness to both semantic drift and structural misalignment. This showcases the strength of our self-similarity guidance, which encourages intra-part consistency and inter-part contrast, enabling meaningful texture placement without explicit mesh supervision. Our flexible guidance based sampling process bridges the gap between structure and high-level semantic intent.

\begin{table}[ht!]
  \centering
  \caption{Quantitative comparison ranking of our \project against baselines for different experimental settings. Results are shown with text conditioning ($\mathcal{L}_\text{structure}$).}
  \vspace{-5pt}
  \resizebox{\linewidth}{!}{
   \begin{tabular}{l|ccc|ccc}
    \toprule
    & \multicolumn{6}{c}{\textbf{Ranking metrics}} \\
      Methods & Fidelity $\downarrow$ & Clarity $\downarrow$ & Adaptation $\downarrow$
      & Fidelity $\downarrow$ & Clarity $\downarrow$ & Adaptation $\downarrow$  \\  
     \midrule\arrayrulecolor{black} 
      & \multicolumn{3}{c}{\cellcolor[HTML]{EEEEEE}{\textit{Intra-Category}}} &
        \multicolumn{3}{c}{\cellcolor[HTML]{EEEEEE}{\textit{Inter-Category}}} \\
        \multicolumn{7}{l}{\cellcolor{colorturquoise}{\textit{Simple-Complex}}} \\
        UV Nearest Neighbor & 3.12 & 3.21 & 3.43 & 3.45 & 3.66 & 3.72 \\
        SDXL + Cross-Image Attention & 2.88 & 2.52 & 3.29 & 3.18 & 2.96 & 3.33 \\
        Trellis~\cite{trellis} & \nd{2.01} & \nd{1.89} & \nd{2.55} & 2.48 & 2.38 & 2.69 \\
        \textit{\project{}} (Ours) & \fs 1.54 & \fs 1.63 & \fs 2.04 & \fs 1.85 & \fs 2.01 & \fs 1.98 \\
    \midrule\arrayrulecolor{black} 
    \multicolumn{7}{l}{\cellcolor{colorturquoise}{\textit{Complex-Complex}}} \\
        UV Nearest Neighbor & 3.45 & 3.52 & 3.61 & 3.58 & 3.67 & 3.75 \\
        SDXL + Cross-Image Attention & 2.97 & 2.78 & 3.22 & 3.15 & 3.06 & 3.31 \\
        Trellis~\cite{trellis} & \nd{2.16} & \nd{2.05} & \nd{2.49} & \nd{2.35} & \nd{2.41} & \nd{2.66}  \\
        \textit{\project{}} (Ours) & \fs 1.36 & \fs 1.48 & \fs 1.53 & \fs 1.72 & \fs 1.87 & \fs 1.90 \\
    \bottomrule
    \end{tabular}
    }
\label{tab:all_cat_text}
\end{table}

\begin{figure*}[h]
    \centering
    \includegraphics[trim=0 0 0 0,clip,width=\linewidth]{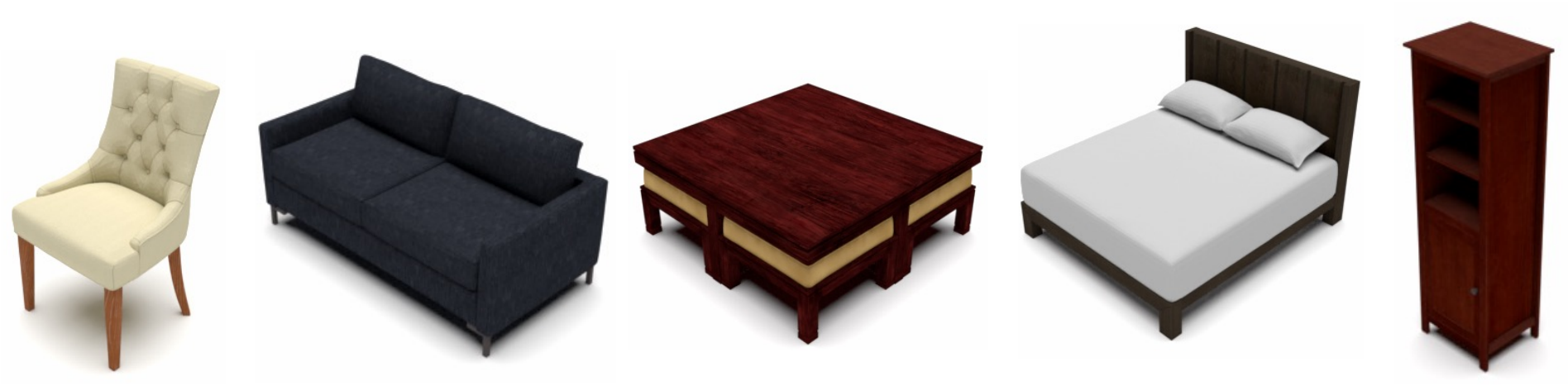}
    \caption{\textbf{Examples of rendered mesh views.}} 
    \label{fig:rendered_mesh_view}
\end{figure*}

\section{Rendered Image Conditioning with $\mathcal{L}_\text{appearance}$}
\label{sec:render_guidance}
In many practical scenarios, an appearance mesh is provided without the corresponding reference image—either because the asset was procedurally generated, sourced from a CAD dataset, or previews were never created. To address this scenario, we render the appearance mesh from a fixed viewpoint to synthesize an image, which serves as the conditioning input for $\mathcal{L}_\text{appearance}$. For all objects we use the same viewpoint, chosen to provide a slightly angled perspective on the object as shown in Fig.~\ref{fig:rendered_mesh_view}. Such view of the appearance mesh captures sufficient visual cues--such as texture, material, and shape context--to serve as effective guidance for our framework. This strategy enables mesh-based guidance in the absence of external visual data. Table~\ref{tab:all_cat_render_meshview} shows results for this setup across intra- and inter-category pairs. Note that these ranking scores are not directly comparable to those in Tab.~\ref{tab:all_cat}, as this experiment includes only the top performing baseline (Trellis); the absence of lower-ranked outputs can shift absolute values due to the relative nature of GPT-based evaluation. While the reference image variant performs best overall, our rendered-view approach still consistently outperforms Trellis\cite{trellis}, demonstrating \project's robustness against limited image data. 

\begin{table}[ht!]
  \centering
  \caption{Quantitative comparison ranking of our \project against baselines for different experimental settings. Results are shown with image conditioning ($\mathcal{L}_\text{appearance}$).}
  \resizebox{\linewidth}{!}{
   \begin{tabular}{l|ccc|ccc}
    \toprule
    & \multicolumn{6}{c}{\textbf{Ranking metrics}} \\
      Methods & Fidelity $\downarrow$ & Clarity $\downarrow$ & Adaptation $\downarrow$
      & Fidelity $\downarrow$ & Clarity $\downarrow$ & Adaptation $\downarrow$  \\  
     \midrule\arrayrulecolor{black} 
      & \multicolumn{3}{c}{\cellcolor[HTML]{EEEEEE}{\textit{Intra-Category}}} &
        \multicolumn{3}{c}{\cellcolor[HTML]{EEEEEE}{\textit{Inter-Category}}} \\
        \multicolumn{7}{l}{\cellcolor{colorturquoise}{\textit{Simple-Complex}}} \\
        Trellis~\cite{trellis} & 2.94 & 2.96 & 2.95 & 2.93 & 2.97 & 2.96 \\
        w/ Rendered Image & 1.96 & 1.98 & 1.99 & 1.97 & 1.99 & 1.98 \\
        w/ Reference Image & \fs 1.08 & \fs 1.03 & \fs 1.02 & \fs 1.06 & \fs 1.02 & \fs 1.01 \\
    \midrule\arrayrulecolor{black} 
    \multicolumn{7}{l}{\cellcolor{colorturquoise}{\textit{Complex-Complex}}} \\
        Trellis~\cite{trellis} & 2.94 & 2.97 & 2.96 & 2.95 & 2.97 & 2.97 \\
        w/ Rendered Image & 1.96 & 1.99 & 1.98 & 1.97 & 1.99 & 1.98 \\
        w/ Reference Image & \fs 1.01 & \fs 1.02 & \fs 1.02 & \fs 1.07 & \fs 1.02 & \fs 1.01 \\
    \bottomrule
    \end{tabular}
    }
\label{tab:all_cat_render_meshview}
\end{table}

\section{Ablation Study}
\label{sec:ablation_study}

We conduct an ablation study to analyze the impact of various design choices in our framework. Results in Tab.~\ref{tab:ablation} show performance on the \textit{simple}-\textit{complex} \textit{intra-category} set under image conditioning. Inspired by the use of global appearance loss in~\cite{tumanyan2022splicing}, (i) we model global appearance for guidance. We do this using a combination of minimum, maximum, and average pooling operations which we found insufficient for our application. This highlights the ineffectiveness of global latents in capturing rich semantic correspondence required for high-quality appearance transfer. (ii) Replacing such global features with nearest-neighbor (NN) matching in \textsc{SLat} space slightly improves performance, especially in fidelity, but lacks robustness in integration and adaptation, confirming that unstructured similarity fails to align content meaningfully across different shapes. (iii) Next, we evaluate a version using K-means-based co-segmentation over \textsc{SLat} features instead of the PartField-driven segmentation~\cite{partfield2025} in \project{}. While guided flow improves generation realism, this variant still underperforms relative to the full method, indicating that semantically informed segmentation is crucial for establishing accurate part correspondences. (iv-v) Finally, we compare the two guidance objectives used in our framework: $\mathcal{L}_\text{appearance}$, and $\mathcal{L}_\text{structure}$, both using image as a condition. Each guidance improves transfer quality in its respective setting, with appearance guidance achieving stronger detail fidelity and structure guidance yielding better alignment and adaptability. These results validate our design of condition-specific guidance, where the choice of loss is tailored to the nature of the input, enabling flexible and semantically consistent transfer.

\begin{table}[ht!]
  \centering
  \caption{Ablation study ranking design choices using image conditioning. Results are on the \textit{simple}-\textit{complex} \textit{intra-category} set.}
  \vspace{-3pt}
  \resizebox{\linewidth}{!}{
   \begin{tabular}{l|cccccc}
    \toprule
    & \multicolumn{6}{c}{\textbf{Ranking metrics}} \\
      Methods & Fidelity $\downarrow$ & Clarity $\downarrow$ & Integration $\downarrow$ & Quality $\downarrow$
      & Adaptation $\downarrow$ & Overall $\downarrow$ \\  
      \toprule
      \multicolumn{7}{l}{\cellcolor{colorturquoise}{\textit{w/o Rectified Flow}}} \\
        (i)  Co-segmentation~\cite{partfield2025} \& global feat. & 4.52 & 4.51 & 4.53 & 4.51 & 4.49 & 4.50 \\
        (ii) Co-segmentation~\cite{partfield2025} \& NN & 3.58 & 3.62 & 3.61 & 3.60 & 3.61 & 3.63 \\
        \midrule\arrayrulecolor{black} 
      \multicolumn{7}{l}{\cellcolor{colorturquoise}{\textit{w/ Guidance in Rectified Flow}}} \\
        (iii)  K-means Co-segmentation + NN & 2.57 & 2.65 & 2.60 & 2.63 & 2.61 & 2.66  \\
        (iv)   w/ Image Condition ($\mathcal{L}_\text{structure}$) - \textit{Ours}  & \underline{2.17} & \underline{2.05} & \underline{2.12} & \underline{2.05} & \underline{2.11} & \underline{2.03} \\
        (v)    w/ Image Condition ($\mathcal{L}_\text{appearance}$) - \textit{Ours} & \textbf{1.23} & \textbf{1.08} & \textbf{1.16} & \textbf{1.11} & \textbf{1.12} & \textbf{1.06} \\
    \bottomrule
    \end{tabular}
    }
\label{tab:ablation}
\end{table}

\section{Application: Scene Editing}
\label{sec:scene_editing}

\begin{figure*}[h]
    \centering
    \includegraphics[trim=0 0 0 0,clip,width=\linewidth]{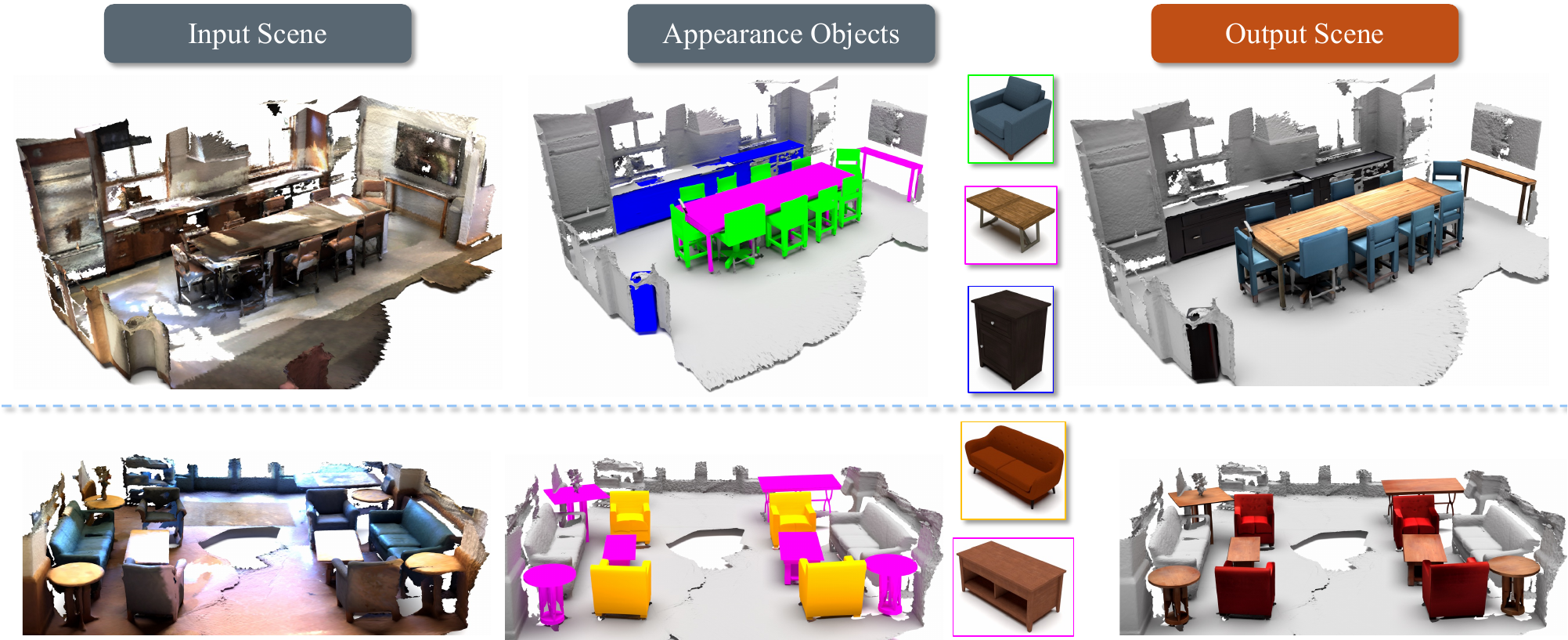}
    \caption{\textbf{Scene editing on scenes from ScanNet~\cite{dai2017scannet}}. For each semantic category, represented with a unique color in the middle column, we select a single appearance object and perform appearance transfer using our \project. Appearance is robustly transferred to different objects in the scene, showing another potential application of our approach.}
    \label{fig:quals_scene_editing}
\end{figure*}

To demonstrate the broader applicability of \project{} beyond isolated object transfer, we explore its use in full-scene editing. In this setting, we start with indoor scans from ScanNet~\cite{dai2017scannet} and utilize per-object CAD mesh annotations from PyTorchGeoNodes~\cite{stekovic2025pytorchgeonodes} and Scannotate~\cite{ainetter2023automatically} to obtain 3D mesh geometries of objects in the scene. For each semantic category present in the scene, we select a representative appearance object from the ABO dataset~\cite{collins2022abo} and perform appearance transfer individually using $\mathcal{L}_\text{appearance}$ on the corresponding meshes. As shown in Fig.~\ref{fig:quals_scene_editing}, our method is capable of stylizing multiple objects across a cluttered scene while preserving their geometric structure. Note that objects with inaccurate pose annotations or high computational demands for Trellis processing are excluded from the transfer process. Despite relying on clean input meshes, our framework generalizes well in complex, real-world environments and opens avenues for future extensions with learned guidance in noisy or incomplete settings. This highlights \project's utility for interactive scene customization, where designers can selectively restyle environments by swapping appearance while preserving structural layout.

\section{Experimental Details}
\label{sec:exp_details}
\noindent \textbf{Evaluation Setup}. For evaluation, we render all assets--including the input, appearance, and output meshes--using Blender with smooth area lighting. Each object is rendered from $4$ viewpoints, sampled around the origin at a fixed radius of $2$ units and a pitch of $30^\circ$, with yaw angles spaced every $90^\circ$ starting from $45^\circ$. All meshes are in a canonical pose to ensure input, appearance reference, and output are aligned, avoiding mismatches like comparing the front side of one object to the back side of another. For the rendered mesh-view guidance experiments (Sec.~\ref{sec:render_guidance}), we use a pitch and yaw of $30^\circ$ and $45^\circ$, respectively, to obtain a single consistent view of the appearance mesh. However, evaluation setup remains the same. Following standard practice~\cite{rottshaham2019singan, cohenbar2025tritexlearningtexturesingle}, we compute every metric, perceptual and GPT-based per view and per object, and report the average across all objects. This ensures that evaluations account for view-dependent variations.

\noindent \textbf{Implementation Details.} Our implementation is based on the pre-trained models and configurations from Trellis~\cite{trellis} and PartField~\cite{partfield2025}. We use \texttt{trellis-image-large} for image conditioning and \texttt{trellis-text-large} for text, and adhere to their default settings for any configuration. For part-aware guidance, we compute part feature fields using PartField~\cite{partfield2025} and query them using the voxel coordinates $p_i$ of each mesh. We run \project{} for single-instance optimization interleaved with rectified flow sampling over 300 steps. Optimization is performed using AdamW with a learning rate of $5 \times 10^{-4}$. All experiments are conducted on a single NVIDIA RTX 4090 GPU. We use identical optimization settings across all conditioning types to ensure fairness and consistency.

\begin{figure*}[t]
    \centering
    \includegraphics[trim=0 0 0 0,clip,width=\linewidth]{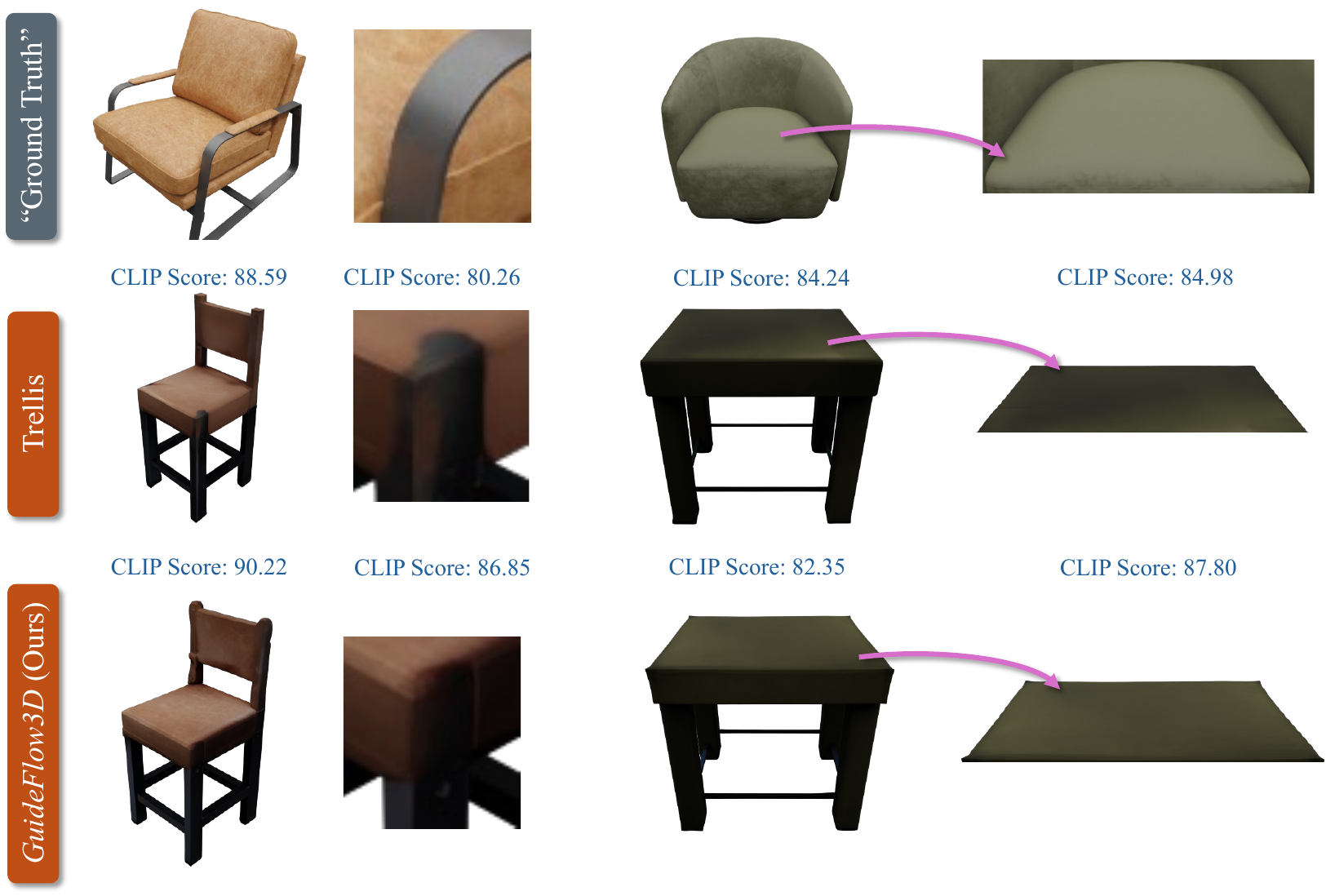}
    \caption{\textbf{Issue with Perceptual Similarity.} Even though our \project visually outperforms the Trellis baseline~\cite{trellis}, CLIP scores do not reflect these improvements. In this example, we observe that the geometric forms of the input and appearance mesh are very different. Therefore CLIP score is not a suitable metric for evaluating similarity of output. We observe that the difference in scores increases for a cropped part of the chair, showing the difference between local and global geometry on the metric.} 
    \label{fig:perceptual_sim_issue}
\end{figure*}

\section{Issues with Perceptual Similarity Evaluation}
\label{sec:perc_sim_eval}

Quantitatively evaluating appearance transfer in 3D is challenging due to the absence of ground truth outputs. Unlike traditional image-to-3D pipelines, where the generated asset can be directly compared to a reference view or textured model, our setting lacks a definitive target for what the transferred appearance "should" look like. As illustrated in Fig.~\ref{fig:perceptual_sim_issue}, the input and output geometries are intentionally dissimilar, and the appearance is adapted--not reconstructed--making traditional reconstruction-based metrics inapplicable. To this end, we report commonly used perceptual similarity scores such as DINOv2~\cite{oquab2023dinov2}, CLIP score~\cite{hessel2021clipscore}, and DreamSim~\cite{fu2023dreamsim} in Table~\ref{tab:same_cat_perceptual}, on image-conditioned appearance transfer for intra-category on \textit{simple}-\textit{complex} set. Due to large geometric differences between appearance and input geometries, these metrics offer only a coarse proxy for evaluating style transfer quality and often fail to capture the localized and semantic nuances essential to this task. Furthermore, in case of conditioning using text prompt, our CLIP score is actually slightly lower than the score for the Trellis baseline which is deceiving. This is because the text prompt often describes a very different geometry which is not aligned with the geometric form of the input object making such metrics less applicable for our experimental setting~(\eg ``padded reclining armchair'' vs. input mesh of a simple stool, as shown in second row of Fig.~\ref{fig:suppl_quals_text}).

\begin{table}[h]
  \centering
  \caption{\textbf{Quantitative comparison based on perceptual similarity metrics.} Results are on the \textit{simple}-\textit{complex} \textit{intra-category} set. Note that DinoV2 and DreamSim can only be evaluated in an image setting.}
  \vspace{-5pt}
  \resizebox{0.6\linewidth}{!}{
   \begin{tabular}{l|cccc}
    \toprule
    & \multicolumn{3}{c}{\textbf{Perceptual similarity}} \\
      Methods & DinoV2 $\uparrow$ & CLIP $\uparrow$ & DreamSim $\downarrow$ \\  
     \midrule\arrayrulecolor{black} 
    \multicolumn{4}{l}{\cellcolor{colorturquoise}{\textit{w/ Image Condition ($\mathcal{L}_\text{appearance}$)}}} \\

        UV Nearest Neighbor & 36.32 & 81.23 & 42.27 \\
        Cross Image Attention & 44.67 & 84.60 & 41.67 \\
        MambaST & 38.33 & 81.73 & 47.55 \\
        EasiTex & 44.42 & 83.80 & 41.13 \\
        Trellis~\cite{trellis} & \nd{52.22} & \nd{87.15} & \nd{36.52} \\
        \textit{\project{}} (Ours) & \fs 52.74 & \fs 87.37 & \fs 36.06 \\

    \midrule\arrayrulecolor{black} 
    \multicolumn{4}{l}{\cellcolor{colorturquoise}{\textit{w/ Text Condition ($\mathcal{L}_\text{structure}$)}}} \\
        UV Nearest Neighbor & - & 25.32 & - \\
        SDXL + Cross-Image Attention & - & 25.96 & - \\
        Trellis~\cite{trellis} & - & \nd{27.64} & \\
        \textit{\project{}} (Ours) & - & \fs 27.23 & - \\
        
    \bottomrule
    \end{tabular}
    }
\label{tab:same_cat_perceptual}
\end{table}

For instance, Fig.~\ref{fig:perceptual_sim_issue} shows an example where our method produces visibly better texture alignment and part-aware fidelity, yet perceptual similarity scores (e.g., CLIP) remain close or even favor inferior outputs. Comparisons on image crops, where scores between methods show larger gaps, highlight how such metrics can overlook structural misalignments, texture stretching, or poor integration when focusing on the global scale that are perceptually obvious to humans. While such encoder-based metrics could be useful in setups with ground-truth part-level correspondences that would enable an evaluation on cropped regions, their effectiveness is limited in our setting where no such reference exists. These limitations underscore the need for our fine-grained, human-aligned evaluation. 

\begin{figure*}[h]
    \centering
    \includegraphics[trim=0 0 0 0,clip,width=\linewidth]{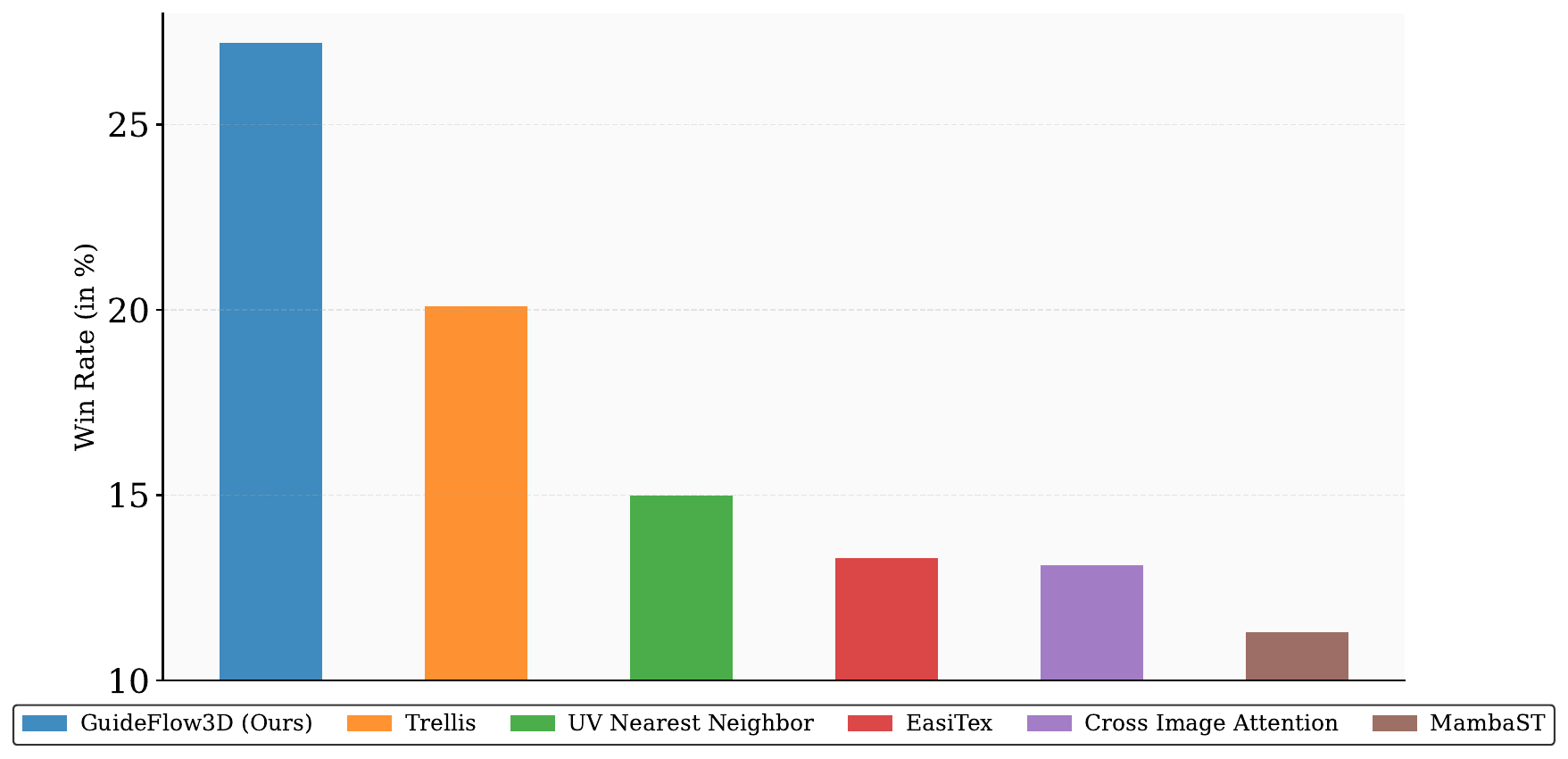}
    \vspace{-15pt}
    \caption{\textbf{Results of Human Evaluation} on randomly sampled outputs from both intra- and inter-category on \textit{simple}-\textit{complex} set.} 
    \label{fig:user_study}
\end{figure*}

\section{Human Evaluation}
\label{sec:user_study}
We conducted a user study on Amazon Mechanical Turk with 59 participants to compare our approach
against baselines. The evaluation set consisted of $100$ randomly selected object and image pairs from the \textit{simple-complex} intra- and inter-category, on the image condition setting. The participants were shown the appearance image and two views (front and back at a camera angle of 45 degrees) each of the structure mesh + outputs from different methods, without any knowledge about the methods. Then, they were provided with the same prompts as the LLM (Figs.~\ref{fig:img_prompt_vis} and~\ref{fig:text_prompt_vis}) and asked to rate the best performing considering overall quality. As shown in Fig.~\ref{fig:user_study}, out of approximately $1000$ evaluations, \project{} received the highest preference ($27.2\%$), demonstrating its ability to balance structural preservation and appearance fidelity. Interestingly, UV Nearest Neighbor was ranked third, despite its lower quantitative performance in Tab.~\ref{tab:main}, likely due to its visually appealing but less semantically grounded results. Overall, the relative ranking across methods remains consistent with our LLM-based evaluation, confirming strong alignment between LLM-based and human judgments of texture fidelity and perceptual quality.

\section{GPT-based Evaluation}
\label{sec:gpt_eval}
\begin{figure*}[ht]
    \centering
    \includegraphics[trim=0 0 0 0,clip,width=\linewidth]{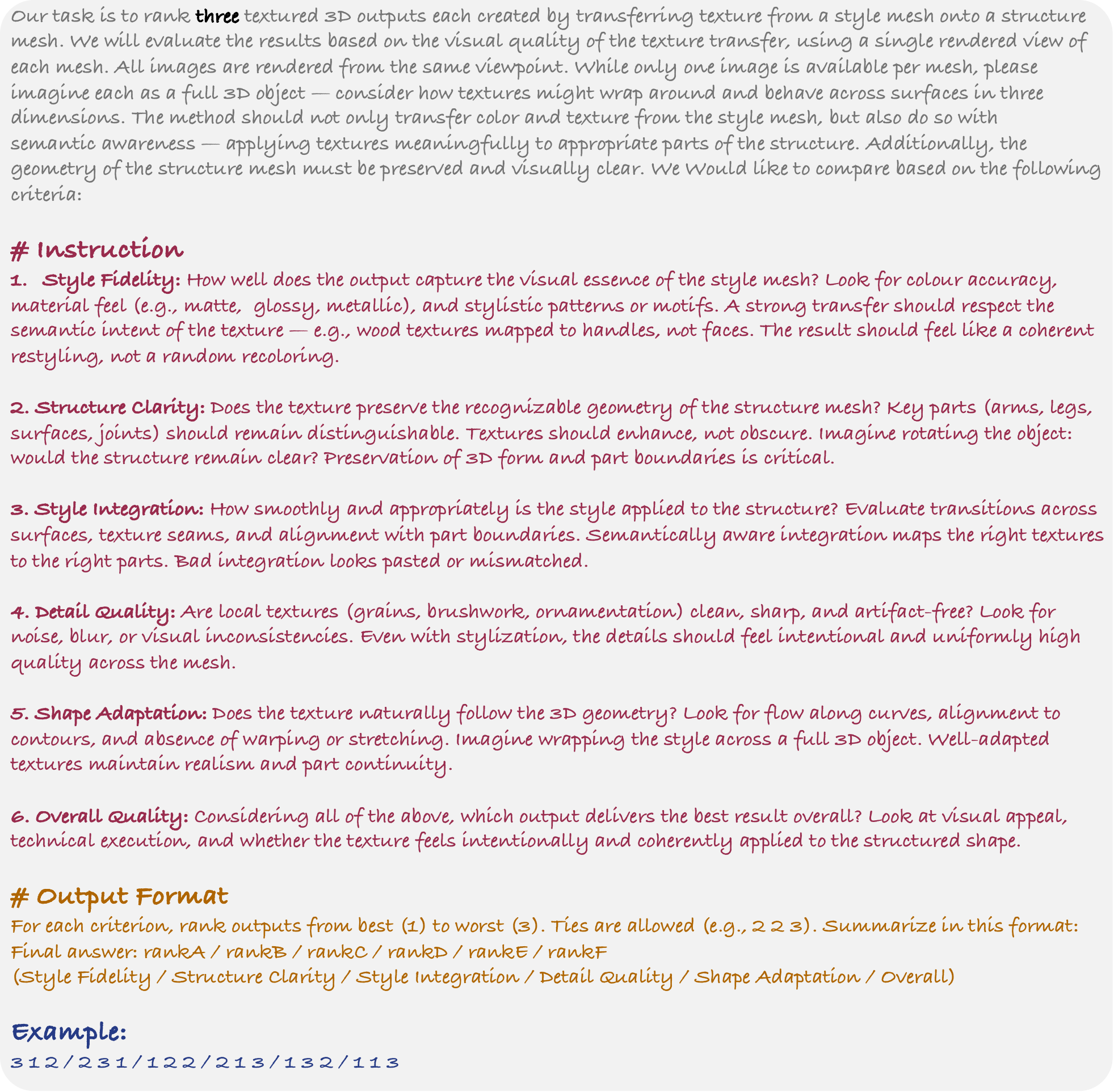}
    \caption{\textbf{Image GPT prompt.} Prompt provided to GPT-5-mini for providing a ranking of textured 3D outputs given an appearance object in the form of an image.} 
    \label{fig:img_prompt_vis}
\end{figure*}

\begin{figure*}[ht]
    \centering
    \includegraphics[trim=0 0 0 0,clip,width=\linewidth]{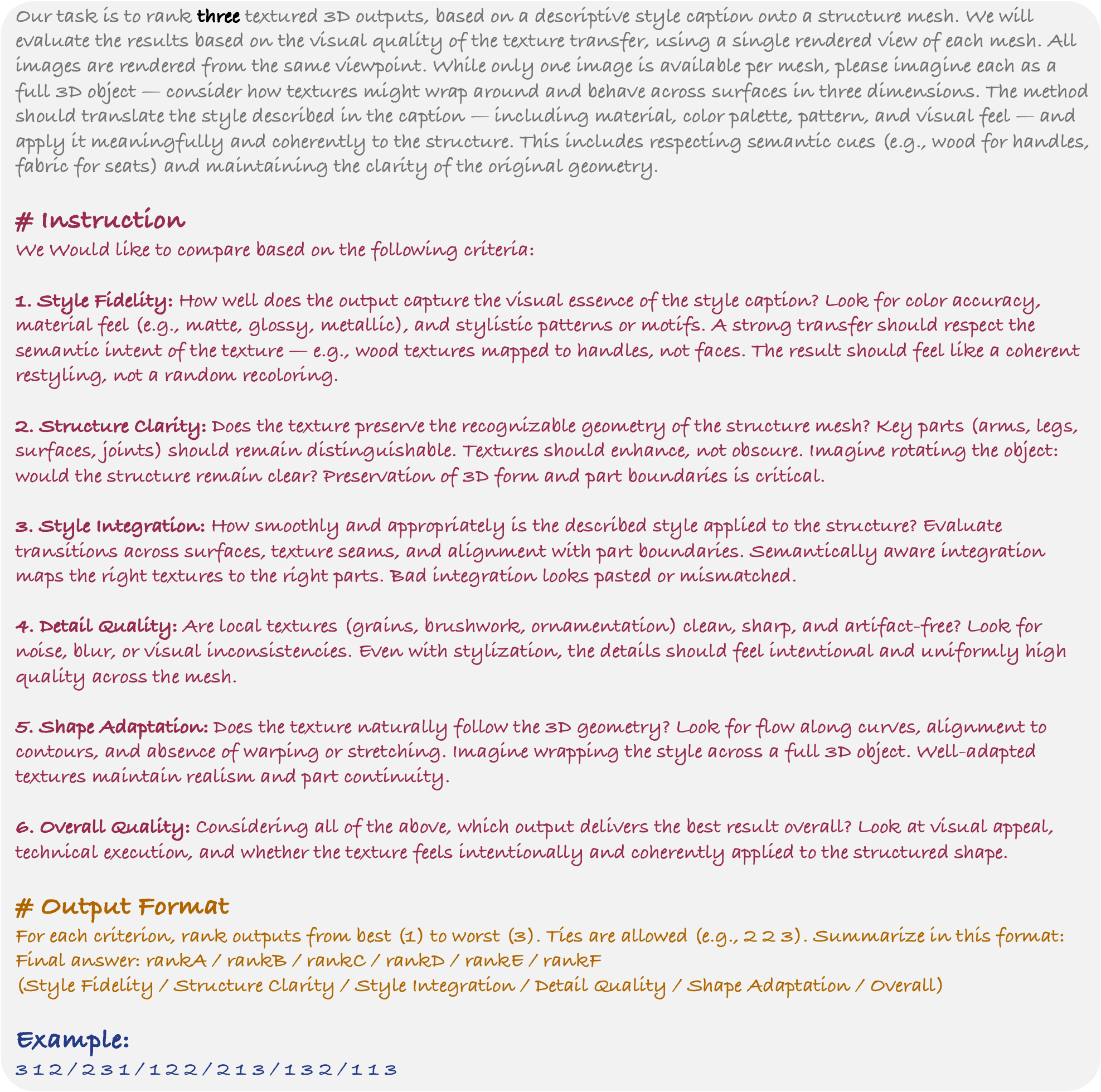}
    \caption{\textbf{Text GPT prompt.} Prompt provided to GPT-5-mini for providing a ranking of textured 3D outputs given an appearance object in the form of a text.} 
    \label{fig:text_prompt_vis}
\end{figure*}

We outline the issues with automated perceptual similarity evaluation in Sec.~\ref{sec:perc_sim_eval} and resort to a GPT-based evaluation mechanism for analysis. GPT-4V has shown strong alignment with human judgments across vision-language tasks, including text-to-3D generation~\cite{peng2024dreambench}, image understanding, and 3D content evaluation.~\cite{wu2023gpteval3d,maiti2025gen3DEval,eval3d2024,Zhang2023GPT4VisionAA}. In designing the GPT-based evaluator, we follow the human-aligned evaluation protocol proposed in GPT-Eval3D~\cite{wu2023gpteval3d} and adopt it for appearance transfer using \textit{gpt-5-mini}. We begin by clearly explaining the task to GPT evaluator: ranking textured 3D outputs based on the quality of appearance transfer from a style source (image or text) to a structure mesh, using a single rendered view per mesh. All images are rendered from a consistent viewpoint, and the evaluator is instructed to imagine each output as a complete 3D object when judging texture consistency and structure preservation. As shown in Fig.~\ref{fig:img_prompt_vis} and Fig.~\ref{fig:text_prompt_vis}, the instruction prompt defines six specific evaluation criteria: Style Fidelity, Structure Clarity, Style Integration, Detail Quality, Shape Adaptation, and Overall Quality. Each criterion is accompanied by a detailed explanation, emphasizing semantic consistency (\eg, appropriately mapping wood textures to structural elements like legs or seats, depending on the design), 3D geometric clarity, texture sharpness, and perceptual coherence. This structured rubric guides the LLM to perform fine-grained comparisons across outputs. To ensure consistency in output formatting and facilitate large-scale evaluation, we explicitly define a fixed output format corresponding to the six criteria. An example format is provided directly in the prompt to reduce ambiguity and improve parsing during automatic result aggregation. Additionally, we support in-context learning by optionally breaking the prompt into multiple stages with example completions shown before evaluation begins. These design choices--structured rubric, visual clarity, strict formatting, and optional in-context examples--ensure that our GPT-based evaluator produces reliable, human-aligned rankings across all experiments. Please note that we use output from all methods for comparison at a time ($\textbf{three}$ is shown as an example in Figs.~\ref{fig:img_prompt_vis} and~\ref{fig:text_prompt_vis}).

\noindent \textbf{Limitations.} While our GPT–based evaluation protocol proved practical and consistent for our purposes, prior work highlights several factors that future users should consider. Large multimodal models are known to exhibit hallucinations~\cite{wu2023gpteval3d, Zhang2023GPT4VisionAA} and systematic biases, such as position sensitivity~\cite{gpt_mt, langsnowball2024}, and preference for high-contrast textures. These factors may distort rankings in ways that do not reflect actual quality. Moreover, GPT-based evaluation could be vulnerable to adversarial artifacts designed to exploit the multimodal model's visual priors. Ensuring such metrics remain `ungamable' is an open challenge. Future research can explore more efficient comparison strategies such as single method evaluation using a scale, \eg $1$ to $5$~\cite{peng2024dreambench}.

\begin{figure*}
    \centering
    \includegraphics[trim=0 0 0 0,clip,width=0.8\textwidth]{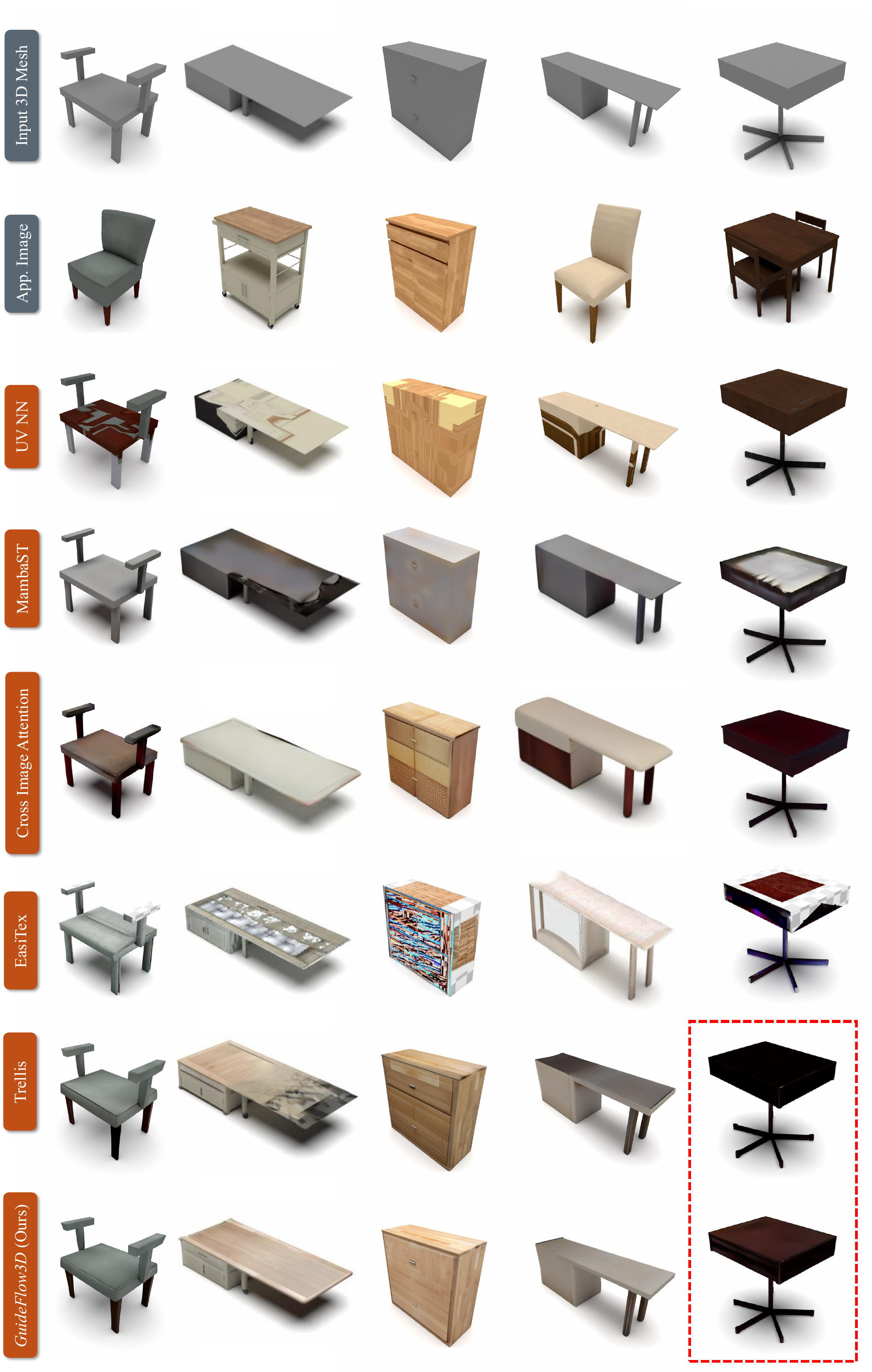}
    \caption{\textbf{Qualitative Comparisons showing results using image condition with $\mathcal{L}_\text{appearance}$.} We show 5 examples (one per column) and the results for all methods. UV Nearest Neighbor (UV NN) often misplaces textures across parts but performs better with uniform materials. MambaST, Cross Image Attention, and EasiTex blend or oversmooth textures, while Trellis improves realism yet lacks geometric grounding. \project{} achieves semantically and structurally consistent transfers, accurately mapping materials like upholstery and wood. Column 5 illustrates a failure case for all methods, though UV NN yields the most visually coherent result and our interleaved guidance scheme cannot recover from the black output produced by Trellis.}
    \label{fig:suppl_quals_image}
\end{figure*}

\section{Qualitative Results}
\label{sec:quals_appendix}
We provide qualitative comparisons for both text- and image-conditioned appearance transfer. Figs.~\ref{fig:suppl_quals_text} and~\ref{fig:suppl_quals_image} highlight \project{}’s ability to produce semantically faithful and geometrically aligned textures across diverse scenarios. In the text-conditioned setting (Fig.~\ref{fig:suppl_quals_text}), we evaluate challenging examples with significant discrepancies between textual descriptions and input geometries. UV Nearest Neighbor and SDXL~\cite{sdxl} + Cross Image Attention~\cite{crossimageattention} produce unrealistic outputs, often failing to capture semantic or color cues. Trellis~\cite{trellis} shows improvements in texture plausibility but struggles with global color consistency and local part correspondence, e.g., missing “reddish-brown” tones or “black metal frames.” In contrast, \project{} effectively grounds textual descriptions in geometry, yielding coherent, part-aware textures that accurately reflect material and structural cues. Nonetheless, when descriptions are abstract (e.g., “metal,” “mirror,” or “cushioned”), all methods show limitations, as shown in Fig.~\ref{fig:suppl_quals_text} (Row 5), highlighting the inherent ambiguity in interpreting such prompts. Under image conditioning (Fig.~\ref{fig:suppl_quals_image}), UV Nearest Neighbor frequently misplaces textures due to limited geometric reasoning, though it performs reasonably when a single uniform texture dominates. MambaST~\cite{mambast}, Cross Image Attention, and EasiTex~\cite{easitex} tend to oversmooth details or blend appearance cues across regions, while Trellis improves realism yet lacks explicit geometric grounding, sometimes introducing artifacts, such as dark textures from correspondence failures. \project{} achieves semantically and structurally consistent transfers, faithfully reproducing upholstery and wood patterns while maintaining sharp boundaries and material coherence. However, certain edge cases remain challenging (Fig.~\ref{fig:suppl_quals_image}, Column 5), such as strong geometry-appearance mismatches or texture flattening under large shape variations. Overall, \project{} demonstrates robust semantic and structural fidelity across both modalities, effectively bridging the gap between appearance semantics and geometric form. These results underscore its capability to generate realistic, production-ready 3D assets.

\end{document}